\documentclass[11pt]{elsarticle}

\usepackage{lineno,hyperref}
\modulolinenumbers[5]

\usepackage{lineno}

\journal{European Journal of Operational Research}
\date{}
\usepackage{amssymb}
\usepackage{mathtools}

\usepackage{graphicx}  
\usepackage{wrapfig}
\usepackage{url}            
\usepackage{booktabs}       
\usepackage{amsfonts}       
\usepackage{nicefrac}       
\usepackage{microtype}      

\usepackage{amsmath}
\usepackage{mathabx}
\usepackage{amsopn}

\usepackage{algpseudocode}
\usepackage{algorithm}

\usepackage{color}
\usepackage{caption}

\usepackage{float}
\floatstyle{plaintop}
\restylefloat{table}

\usepackage{subfigure}
\usepackage{amsthm}
\usepackage{bm}

\DeclareMathOperator*{\argmax}{arg\,max}
\DeclareMathOperator*{\argmin}{arg\,min}

\newtheorem{definition}{Definition}
\newtheorem{lemma}{Lemma}

\usepackage{color,soul}
\usepackage[usestackEOL]{stackengine}[2013-09-11]

\setstackgap{L}{1.4\baselineskip}
\iftrue
\let\oldthebibliography\thebibliography
\let\endoldthebibliography\endthebibliography
\renewenvironment{thebibliography}[1]{%
  \oldthebibliography{#1}%
  \setlength{\itemsep}{-1pt}%
  \setlength{\parskip}{0pt}%
  \linespread{0.9}\selectfont
}{\endoldthebibliography}
\fi

\biboptions{authoryear}
\begin{document}

\begin{frontmatter}

\title{Opponent Aware Reinforcement Learning}


\author[mymainaddress,komorebiaddress]{Víctor Gallego}
\cortext[mycorrespondingauthor]{Corresponding author: victor.gallego@komorebi.ai}

\author[cunefaddress]{Roi Naveiro}

\author[mymainaddress]{David Ríos Insua}

\author[mysecondaryaddress]{David Gómez-Ullate}

\address[mymainaddress]{Institute of Mathematical Sciences (ICMAT), Madrid, Spain.}
\address[komorebiaddress]{Komorebi AI Technologies, Madrid, Spain.}
\address[cunefaddress]{CUNEF Universidad, Madrid, Spain.}
\address[mysecondaryaddress]{School of Science and Technology, IE University, Madrid, Spain.}

\begin{abstract}
In certain reinforcement learning (RL) scenarios
there are adversaries trying to \textcolor{black}{interfere} with the underlying reward process
\textcolor{black}{ for their own benefit}.
We introduce \textcolor{black}{{\em Threatened Markov Decision Processes}} (TMDPs) 
as  a framework to support an agent against potential opponents
in an RL context as well as schemes 
resulting in novel learning approaches to deal with TMDPs. After introducing
our framework and deriving theoretical results, empirical evidence is given via extensive experiments, showing the importance 
\textcolor{black}{ for an RL agent of acknowledging adversarial awareness.} 
 
\end{abstract}


\end{frontmatter}


\section{Introduction}
Over the last decade, an increasing number of processes are being automated through 
machine learning (ML) algorithms \textcolor{black}{ as illustrated in  e.g. \textcolor{black}{\citet{DEEPLEARNING}}} or
\citet{murphy2023probabilistic}. {\color{black} It is therefore essential that these algorithms are robust and reliable so that operations based on their output can be trusted}. State-of-the-art
ML algorithms perform extraordinarily well on standard data. \textcolor{black}{However, they   
have been repeatedly shown to be vulnerable to, \textcolor{black}{ e.g., adversarial examples} 
\citep{goodfellow2014explaining} \textcolor{black}{ purposefully designed to fool ML algorithms}.}

\textcolor{black}{ The pioneering work on adversarial classification (AC) 
by \citet{dalvi2004adversarial} presented the confrontation between a learning system and adversaries through the lens of   
 game theory \citep{menache2011network} 
 largely framing the field of adversarial machine learning (AML),}
 see reviews in \textcolor{black}{\citet{BIGGIO2018317},  \citet{doi:10.1002/widm.1259}, and \citet{jasa}}. This entails  
 well-known common
knowledge assumptions, {\color{black} critically analyzed in e.g.\ \citet{hargreaves2004game}.
\textcolor{black}{ From
a fundamental point of view, such assumptions are not sustainable in 
security settings 
where adversaries tend to 
conceal information}. 
In recent work \citep{naveiro2018adversarial,gallego2024protecting}, we presented novel 
frameworks for AC based on Adversarial Risk Analysis (ARA) {\color{black}\citep{roponen}}.
\textcolor{black}{ ARA is a decision-theoretic approach to games \citep{hausken2024fifty} providing} one-sided prescriptive
support to an expected utility maximizing agent
 making operational the Bayesian approach to games in {\color{black}\citet{kadane1982subjective} and \citet{raiffa1982art}}, by facilitating procedures to
predict adversarial decisions. 

The AML literature has predominantly focused on
the supervised learning setup.  
Our focus will be in reinforcement learning (RL) {\color{black}\citep{sutton2012reinforcement,powell2022reinforcement}}.
In it, an agent takes actions
sequentially to maximize \textcolor{black}{an expected cumulative reward}, learning from
interactions with the environment \citep{kaelbling1996reinforcement}. \textcolor{black}{ With the advent of deep learning \citep{gallego2022current},} deep RL
has experienced an incredible growth
\citep{mnih2015human,silver2017mastering} with important 
applications in domains such as automated driving systems or autonomous robots\textcolor{black}{, as well as core operational research problems such as inventory control \citep{boute2022deep}}.
However, RL systems may be also the target of 
adversarial attacks \citep{huang2017adversarial,lin2017tactics} 
and robust learning methods are thus in need. 
A related field of interest is multi-agent RL (MARL) \citep{leelee},
where multiple 
agents try to learn to compete or cooperate. 
 Importantly, single-agent RL methods
tend to fail in
these settings since they do not take into account 
the non-stationarity   
 \textcolor{black}{  due to the actions and learning of the other agents present in the RL
 scenario, as \citet{BusoniuMARL,marl_over} argue}. 

\textcolor{black}{ Based on the note by \citet{gallego2019reinforcement}},
 \textcolor{black}{  our major contribution here is
to demonstrate how ARA facilitates dealing with 
secure RL, developing
an approach to model and predict adversarial actions aimed at interfering}  with the reward
generating process. 
\textcolor{black}{Unlike earlier MARL approaches, which typically focus on specific aspects of learning, such as worst-case scenarios or stationary opponent models, we present a global framework and provide extensive empirical evidence of its efficiency, flexibility, and robustness.}
  \textcolor{black}{ We deliver strategies to predict opponents' policies},   
 depending on the information and computational resources available: }
  \textcolor{black}{  an approach inspired by fictitious play,} 
  a scheme based on level-$k$ thinking, and a model averaging procedure
  combining several adversarial models. Importantly, we extend 
the approach to deep RL scenarios.
A variety of security experiments 
  \textcolor{black}{ illustrate  the 
generality of our proposal, covering issues like the advantages of 
  incorporating accurate adversarial prediction models, 
  the robustness
in \textcolor{black}{the }presence of misspecified models, and how to handle multiple opponents.}

 \textcolor{black}{ To facilitate reproducibility, all the code and experimental setup details
are released at \url{https://github.com/vicgalle/ARAMARL}.}


\section{Background and Related Work}\label{sec:background}

\textcolor{black}{Our focus is on RL, a well-studied computational approach addressing \textit{Markov decision processes} (MDPs) \citep{howard:dp}: a single agent (the decision maker, or DM) makes decisions while interacting with an environment over discrete time steps. An MDP is represented by the tuple $\left( \mathcal{S}, \mathcal{A}, \mathcal{T}, R\right)$, described as follows.\footnote{In its definition, $\Delta(X)$ 
 is used to designate the set of all probability distributions over $X$, endowed with an appropriate  probabilistic structure}  At each time step, the agent makes decisions $a_t$ from a feasible set $\mathcal{A}$ within an environment $\mathcal{S}$ whose states are $s$. Such decision leads to a new random state $s_{t+1}$ characterized by a function
$\mathcal{T}: \mathcal{S} \times \mathcal{A} \rightarrow \Delta (\mathcal{S})$ modeling the transition distribution associated with a state $s$ and action 
$a$. 
   Besides, it also provides the agent with a random 
$R : \mathcal{S} \times \mathcal{A}  \rightarrow \Delta(\mathbb{R}) $, 
  where $R(s_t,a_t)$ is the distribution 
over rewards perceived from state $s_t$ and action $a_t$, with 
realizations $r(s_t,a_t)$.} 
The DM chooses her actions based on a policy $\pi: \mathcal{S} \rightarrow \Delta(\mathcal{A})$, indicating for each state $s$ the distribution from which 
an action $a$ is drawn.  Her aim is to maximize  
her long-term discounted expected reward, that is, 
$ 
 \mathbb{E}_{\tau} \left[  \sum_{t=0}^\infty \gamma^t R(s_t, a_t) \right]{\color{black},} 
 $ 
where $\gamma \in (0,1)$ is a discount factor and $\tau = (s_0, a_0, s_1, a_1, \ldots)$ is a trajectory of states and actions.

 \textcolor{black}{ Our discussion will mainly center around $Q$-learning 
 \citep{sutton2012reinforcement} an efficient approach in numerous 
 RL settings, but see our final conclusions for extensions}. \textcolor{black}{ In it, the DM maintains a
 function
$Q : \mathcal{S} \times \mathcal{A} \rightarrow \mathbb{R}$ that estimates
her expected cumulative reward, updated according to}
\begin{align}\label{eq:ql}
Q(s,a) &:= (1 - \alpha) \, Q(s, a)  +  \alpha \, \left(r(s,a) + \gamma \, \max_{a'} Q(s', a') \right),
\end{align}
where $\alpha$ is a learning rate hyperparameter
and \textcolor{black}{ $s'$ designates the state 
after having chosen action $a$ from state $s$.} 
\textcolor{black}{We focus on $Q$-learning for four reasons: (i) its off-policy nature lets the DM bootstrap from observed transitions, naturally absorbing the opponent's actions and rewards; (ii) the $\max$ in~(\ref{eq:ql}) admits a clean extension by averaging the bootstrapped value over the DM's beliefs about the opponent's move, \textcolor{black}{ the technical core of this paper
expressed through rule~(\ref{eq:lr})-(\ref{eq:lr2})}; (iii) that rule preserves Bellman contractive structure, so its 
 convergence analysis (\ref{sec:p}) carries over with minor modifications; and, (iv) the machinery is agnostic to tabular vs.\ parametric/deep representations of $Q$ (Section~\ref{sec:approx}). Stochastic policy-based methods such as policy gradients \citep{baxter2000direct}, trusted region policy optimization (TRPO) \citep{pmlr-v37-schulman15}, proximal policy optimization (PPO) \citep{schulman2017proximal} and actor-critic \citep{konda2000actor} are equally applicable, and may be preferable when the equilibrium is mixed (e.g., matching-pennies-like settings); the policy-gradient surrogate is then taken in expectation over $p_A(b\mid s)$, and the level-$k$ and type-based schemes of Sections~\ref{sec:k}--\ref{sec:com} can wrap any base learner that maintains an estimate of $p_A$. We elaborate this in Sections~\ref{sec:discussion} and 5.}

\textcolor{black}{ Our major interest is in security settings where \textcolor{black}{ another  agent
interferes} with the DM's reward process \textcolor{black}{ for their own benefit}. As an example,
\textcolor{black}{ in an air combat scenario,
the rewards of a supported pilot 
   are affected by the performance} 
of an enemy pilot.
This renders the environment
non-stationary, in particular making $Q$-learning suboptimal \citep{marl_over}.}
Thus, to support the agent,
we must be able to reason about and forecast the adversary's behavior.
\textcolor{black}{ Since our concern is on the ability of our agent to predict 
her adversary's actions, we focus on methods
with this goal, encompassing three of the 
  broad opponent modeling principles proposed in the  literature,
  see e.g.\ {\color{black} by} \citet{Albrecht2018AutonomousAM}.} 
\textcolor{black}{  First, {\em policy reconstruction} methods aim to 
fully reconstruct} the adversary's decision
making problem, typically assuming a parametric model fitted  
after observing the adversary's behavior. A representative approach 
stems from fictitious play \citep[FP,][]{brown1951iterative} modeling  
the other agents by computing their frequencies of choosing various actions.
 Next, 
{\em type-based reasoning} assume{\color{black}s} that the modeled agent belongs
to one of several fully specified types and learns 
a distribution over such types, 
without explicitly including the ability of
other agents to reason about their opponents' decision making. 
Finally, \textcolor{black}{\em recursive learning} methods 
stem from explicit representations of the other agents' beliefs about their
opponents leading to an infinite hierarchy of decision making problems,
  illustrated {\color{black}by} \citet{rios2012adversarial} in a simpler class of problems. 

\textcolor{black}{These opponent modeling principles have been developed mostly within the decision analysis arena  \citep{banks2022adversarial}, and their integration with $Q$-learning in multi-agent RL settings remains, to our knowledge, largely unexplored. To articulate the gap that our work fills, we organize the related literature into four threads.}
\vspace{-0.05in}

\paragraph{\textcolor{black}{Multi-agent $Q$-learning via Markov games}} \textcolor{black}{Extensions of $Q$-learning to multi-agent settings have focused on modeling the whole system through Markov games.  Three well-known solutions 
 are  {\em minimax-$Q$ learning} \citep{littman1994markov}, where at each iteration a minimax
problem is solved; {\em Nash-$Q$ learning} \citep{hu2003nash},
  generalizing  the previous algorithm to non-zero sum scenarios;
and {\em friend-or-foe-$Q$ learning} \citep{littman2001friend}, in which the DM knows in advance whether her opponent is an adversary or a collaborator. Within the bandit literature, \citet{auer1995gambling} introduced a non-stationary setting where the reward process is affected by an adversary. All of these methods inherit
standard game-theoretic common-knowledge assumptions \citep{hargreaves2004game}, unrealistic in the security domains of interest to us.} Our approach departs fundamentally from this 
work in that we consider the problem from a single DM's perspective and
explicitly model the opponent for which we propose several strategies
leading to important operational advantages. 

\paragraph{\textcolor{black}{Opponent modeling in deep RL}} \textcolor{black}{A second thread incorporates explicit opponent models into deep RL.} Our work relates to
\citet{lanctot2017unified}, who propose a deep
cognitive hierarchy as an approximation to a response oracle
to obtain policies that can exploit adversaries. Ours
rather draws on the
level-$k$ thinking tradition \citep{stahl1994experimental}, building opponent models that support the DM in predicting
their behavior. In addition, we provide a solution to
the important issue
of choosing between different opponent cognition levels. \textcolor{black}{Closely related is neural fictitious self-play \citep{heinrich2016deep} and the broader self-play family that \citet{lanctot2017unified} encompass\textcolor{black}{. These} methods extend FP to deep RL by approximating best responses to the empirical average policy of past play, with the explicit goal of converging to a Nash equilibrium of the underlying game. Our framework departs from this equilibrium-seeking stance: TMDPs provide Bayesian prescriptive support to a single DM, treating \textcolor{black}{ the opponents' behavior as sources} of uncertainty whose policy is predicted from its posterior rather than averaged into a self-play target. This makes the framework adaptive under non-stationarity and model misspecification, since the DM continually updates her beliefs about the opponent, instead of relying on the convergence of a joint best-response dynamic.}
\citet{he2016opponent} also addressed opponent modeling
in deep RL scenarios. However, the authors rely on using a specific neural network (NN) architecture over the deep $Q$-network model. Instead, we tackle the problem without assuming a NN architecture for the agent policies, though our proposed scheme can be adapted to that setting, as Section 3.4 shows. \citet{foerster2018learning}
adopt a similar experimental setting {\textcolor{black}{in their LOLA framework}}, but their methods apply only
\textcolor{black}{when both players have exact knowledge of their opponents' policy parameters or a maximum-likelihood estimator,  still a form of common knowledge.} Our work
instead builds upon estimating the opponents' $Q$-function, not requiring direct knowledge of the opponent's parameters. \citet{wen2019probabilistic} use level-$k$ thinking in deep RL, yet they only go up to level 2 and their framework
does not extend easily to higher levels, unlike ours which
is generic. Even at the same cognitive level, their framework is more expensive computationally, as they rely on
variational inference to estimate several latent variables.
\textcolor{black}{More recent developments cluster around two core ideas. \emph{Opponent-shaping} methods directly succeeding LOLA steer the opponent's learning process: POLA \citep{zhao2022proximal} makes LOLA updates proximal but still requires the opponent's policy parameters; M-FOS \citep{lu2022model} instead meta-learns a shaping policy over many opponent lifetimes; LOQA \citep{aghajohari2024loqa} shapes opponents modeled as softmax $Q$-learners; and advantage alignment \citep{duque2025advantage} unifies this family. \emph{Opponent-modeling} methods include MBOM \citep{yu2022model}, a deep counterpart of the level-$k$ hierarchy with type-based averaging developed below, LIAM \citep{papoudakis2021agent}, and GSCU \citep{fu2022greedy}. Section~\ref{sec:discussion} discusses their informational assumptions  
 and we compare our approach against LOQA, POLA, and MBOM in Section~\ref{sec:loqa}.}

\paragraph{\textcolor{black}{Adversary-aware planning frameworks beyond RL}} \textcolor{black}{A third thread tackles adversarial decision making outside the standard RL setting.} These include links with iPOMDPs as in \citet{gmytrasiewicz2005framework} and, more broadly, with Partially Observable Stochastic Games (POSGs) \citep{hansen2004dynamic}\textcolor{black}{. These} authors
refer only to planning,
\textcolor{black}{not addressing the typical RL scenario where the DM has to learn as well from the environment}. Moreover, both formalisms are notoriously intractable: POSG planning is NEXP-complete, whereas i-POMDPs require maintaining beliefs over an, in principle, infinite hierarchy of nested beliefs, only computationally feasible after aggressive truncation. Our level-$k$ bounded-rationality scheme can be viewed as a deliberate, finitely truncated counterpart of the i-POMDP nested-belief regress, designed to preserve computations linear in $k$, while still supporting Bayesian updating about the opponent. In addition, we introduce a more simplified algorithmic apparatus (in terms of both implementation and complexity) that performs well in examples and covers the case of mixtures of opponent types.
\citet{caballero2021identifying} \textcolor{black}{ground their approach on level-$k$ thinking to compute optimal strategies in normal form games, relying on solving a stochastic optimization program exactly.} As our framework also tackles MDPs, their approach would be infeasible in this case due to the large number of states. Therefore, \textcolor{black}{ we resort to our upgraded Q-learning scheme}.

\paragraph{\textcolor{black}{Robustness and adversarial attacks on RL}} \textcolor{black}{A fourth thread targets the robustness of RL agents to specific attack vectors rather than strategic opponent modeling.} \citet{pinto2017robust} propose a method for deep RL based on policy gradients
which works only for zero-sum games. In addition, their experiments focus on single-agent RL settings not accounting for an adversary.
  \citet{banihashem2021defense} focus on poisoning attacks over the reward process of an offline learning agent, another paradigm to deal with RL.  \citet{caballero2021poisoning} also deal with poisoning attacks, but their work is restricted to MDPs with known transition distributions, there being no learning of the environment nor the adversaries. There is also a fair amount of work focusing on adversarial perturbations of the (visual) inputs of the learning agent, without strategically modeling the adversary who could also be learning, as in our framework\textcolor{black}{. Some references are }   \citet{behzadan2017vulnerability,zhang2020robust,chen2019adversarial},
and \citet{tretschk2018sequential}.

\textcolor{black}{In summary, across these four threads, the literature either assumes common knowledge among agents, restricts opponent reasoning to specific NN parameterizations or fixed cognitive depths, addresses planning rather than learning, or targets robustness against specific attack vectors instead of strategic opponent modeling. The gap we fill is the integration of the three opponent modeling principles mentioned earlier (policy reconstruction, type-based reasoning, and recursive learning) into (mainly) $Q$-learning, through the single-DM perspective advocated by ARA \citep{rios2009adversarial}.} 

To this end, consider the problem of
prescribing decisions to a single agent versus her opponents,
augmenting the MDP to account for potential adversaries  conveniently
re-defining the $Q$-learning rule. This enables us to apply some of the previously mentioned modeling techniques
to the $Q$-learning setting, explicitly accounting
for the possible lack of information about the modeled opponent.
Our focus will be on the case of a DM (agent $A$, she) facing a single opponent ($B$, he), although Section \ref{sec:mul} provides an extension to a setting with multiple adversaries illustrated in
 Section 4.2.

\section{Threatened Markov Decision Processes}
 \textcolor{black}{ 
 \textcolor{black}{ Our concept of {\em Threatened Markov Decision Process}} (TMDP) addresses the presence of an adversary who modifies state and reward dynamics resulting in a non-stationary environment. This MDP augmentation considers modified transition and reward processes and incorporates the agent's beliefs about the adversary's actions.
We thus adopt a Bayesian approach to games, in the  
\citet{kadane1982subjective} and \citet{raiffa1982art} sense, and \textcolor{black}{ making it 
 operational in } multi-stage
 contexts.}
\begin{definition}
\textcolor{black}{A 
 TMDP is a tuple
$\left( \mathcal{S}, \mathcal{A}, \mathcal{B}, \mathcal{T}, R, p_A \right) $
in which $\mathcal{S}$ is the space of states $s$; $\mathcal{A}$ is the set of
actions $a$ available to the supported agent $A$; $\mathcal{B}$ is the set of
 threat actions $b$ 
available to the adversary $B$; 
$\mathcal{T}: \mathcal{S} \times \mathcal{A} 
\times \mathcal{B} \rightarrow \Delta(\mathcal{S})$ is the {\color{black}(possibly unknown)}
transition distribution; 
$R : \mathcal{S} \times \mathcal{A} \times \mathcal{B} \rightarrow
\Delta(\mathbb{R}) $ is the distribution 
over the reward that 
 agent $A$ perceives from a given state $s$ when the pair of actions
$(a,b)$ is implemented; and $p_A (b | s)$ represents the
 distribution 
   modeling 
 $A$'s beliefs about
 $B$'s moves, when the current state is $s \in \mathcal{S}$.}
\end{definition}

\textcolor{black}{ To ground Definition~1 in a realistic setting, consider beyond-visual-range (BVR) air-combat decision support, in the spirit of \citet{virtanen2006modeling} and  \citet{camacho2024algorithmic}.
The supported agent~$A$ is a blue pilot. Her adversary~$B$, a red one. The state $s$ aggregates the kinematic variables of both aircraft (position, velocity, heading, pitch and roll), the remaining missile inventories, and indicators such as whether a missile is currently in flight or the target is within radar. The DM's action set $\mathcal{A}$ specifies whether to launch a missile and which waypoint (heading) to pursue, while the adversary's action set $\mathcal{B}$ comprises canonical tactical maneuvers like \emph{search}, \emph{launch-and-support}, \emph{pump}, \emph{recommit}, or \emph{withdraw}. 
The transition $\mathcal{T}$ is induced by a flight-dynamics simulator together with a stochastic weapon-engagement server. Reward $R$ combines a missile-launch term peaked at the optimal firing range, weapon-engagement-zone shaping, and a penalty for entering $B$'s minimum-abort range. The    forecast $p_A(b\mid s)$ is then naturally a posterior over the adversary's latent risk profile, updated from the post-engagement observability of his maneuvering that is routine in after-action reports and tactical replays. The  machinery developed below would let the blue pilot anticipate that her opponent may himself be reasoning about her launches and waypoints. This is precisely the prescriptive lift TMDPs provide over an opponent-unaware controller.}

\color{black}{\paragraph{Note 1} This definition of a TMDP is relative to the supported
agent $A$. \textcolor{black}{ In particular, notation $p_A$ remarks that $A$'s
 beliefs about $B$'s actions are being modeled.} When  
 supporting the other agent, we would have to model her corresponding TMDP with her rewards and beliefs. Similarly, the concept and notation extends when there is more than one adversary, as Sections 3.5 and 4.2 illustrate. \hfill $\triangle$
{\color{black}\paragraph{Note 2}
Common conventions in the RL and stochastic optimization communities recommend including all necessary information in the system's state variables to compute the policy, transitions, or cumulative rewards \textcolor{black}{(see, e.g., \citealp{powell2022reinforcement}\textcolor{black}{; \citealp{powell2019unified}})}. In our case, this would mean incorporating the opponent's actions into the state. However,    while the literature in these fields typically assumes a stationary environment, 
 our setting is different: other agents are also learning, causing their policies, and, consequently, $A$'s estimate of $p_A(b | s)$, to evolve over time, making the environment non-stationary. \textcolor{black}{ In TMDPs, the non-stationary component ($A$'s beliefs about $B$'s actions) is treated separately from the stationary component. 
   This simplifies the notation and allows us to explicitly describe how non-stationary elements are handled.  \hfill $\triangle$}
}
\vspace{0.1in}

 {\color{black} TMDPs will be viewed as a relevant extension to MDPs with discounted rewards.
To deal with TMDPs, we modify the standard $Q$-learning update rule (\ref{eq:ql}) 
by forecasting $B$'s actions 
 and averaging over them. This way the DM anticipates potential threats within her
decision making process, enhancing the robustness of her decision-making policy. Formally,}   replace the 
  update rule (\ref{eq:ql})}  \textcolor{black}{ by 
\begin{align}\label{eq:lr}
Q(s, a, b) := (1 - \alpha)Q(s, a, b) +  \alpha \left( r(s,a,b)
+ \gamma \max_{a'} \mathbb{E}_{p_A(b|s')} \left[ Q(s',a',b)  \right]  \right){\color{black},}
\end{align}
where $s'$ is the state reached after the DM and her adversary adopt, respectively,
actions $a$ and $b$ at state  $s$. Then, compute the 
expectation of $Q(s,a,b)$ taking into account the
uncertainty about  the opponent's action 
\begin{align}\label{eq:lr2}
Q(s,a) := \mathbb{E}_{p_A(b|s)} \left[ Q(s,a,b) \right]{\color{black}.} 
\end{align}}
 \ref{sec:p} proves the convergence
of (2)-(3). 

This result only applies when the adversary's policy is stationary, i.e., its distribution does not change over time. \textcolor{black}{  However, as mentioned, in realistic scenarios } the adversary may also learn from our actions, so that his policy will change over time. To deal with this, the next subsections will propose several 
 approaches to model adversarial behavior.
 Furthermore, \textcolor{black}{ in most of our experiments we use $Q(s,a)$ to compute an $\epsilon$-greedy policy for the DM when the system is at state $s$: with probability $(1-\epsilon)$ choose $a^* = \argmax_a  \left[ Q(s,a)  \right] $; with probability $\epsilon$,
choose a random action uniformly.  Yet  other sampling methods such as softmax can be straightforwardly used
as Section 4.1.1 illustrates.}

\begin{figure}[hbt]
   \centering
   \includegraphics[width=0.85\linewidth]{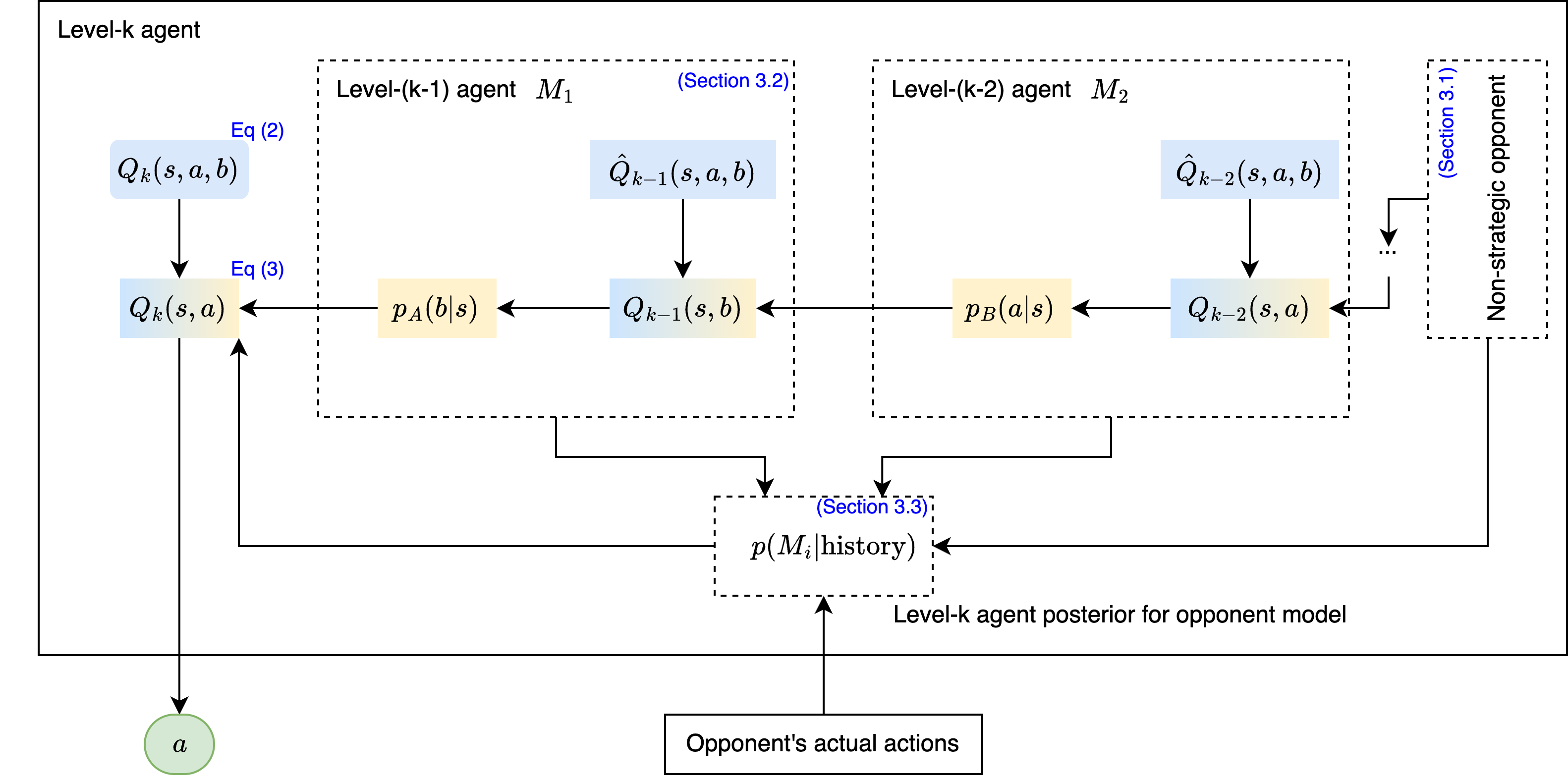}
   \caption{\textcolor{black}{Framework diagram. A level-$k$ agent internally models her opponent as a level-$(k-1)$ agent until arriving at a non-strategic opponent. The level-$k$ agent has to estimate the $Q$-value tables (blue), and extract the corresponding policy for each of the adversarial models (yellow). Optionally, the level-$k$ agent may place a prior over the opponent model distribution and update it. All $Q$-value tables are updated within each environment update, as the agent perceives her opponent(s) decisions.}} \label{fig:oarl_diagram}
 \end{figure}

\textcolor{black} { Since common knowledge is not imposed,
 the core difficulty lies in modeling agent $A$'s uncertainty regarding 
adversary $B$'s policy through  $p_A (b | s)$.
 For this, we make the 
standard multi-agent RL assumption that both agents observe their opponent's actions and rewards
after they have committed to them.  Based on it, three approaches 
are proposed to forecast the opponent's policy $p_A (b | s)$,
depending on the information and computational resources available. 
  In the first one, the adversary is assumed to be non-strategic, acting 
 without awareness of the DM, and an FP inspired forecast 
  is presented.  Secondly, more sophisticated adversaries
  are modeled with a scheme based on level-$k$ thinking. Finally,
   a mixture method combining different opponent models is outlined, allowing us to deal with mixed behavior. Figure \ref{fig:oarl_diagram}
    presents a vignette with these components. 
    We complete the discussion suggesting how to deal with multiple opponents,
 an extension to deep RL scenarios, and present 
 computational complexity issues.}

\textcolor{black}{ 
\subsection{Non-strategic (level-0) opponents}\label{sec:non}
 \paragraph{Stateless setting} To fix ideas, consider  first a stateless setting with discrete action spaces $\mathcal{A}$ and $\mathcal{B}$.
 The $Q$-function (\ref{eq:lr}) is displayed \textcolor{black}{ as a table} 
$Q(a_i,b_j)$, with $a_i \in \mathcal{A}$ the action chosen by
the DM and $b_j \in \mathcal{B}$ the action chosen by the adversary.  
Then, the DM  computes the expected reward of action $a_i$ using the stateless version of \eqref{eq:lr2},
 $ \psi(a_i) = \mathbb{E}_{p_A(b)} [Q(a,b)] = \sum_{b_j \in \mathcal{B}} Q(a_i, b_j) p_{A}(b_j),$ 
where $p_A (b_j)$ reflects $A$'s beliefs about her opponent's actions.
Finally, she chooses the action $a_i \in \mathcal{A}$
maximizing $\psi(a_i)$.}
 
  Our first option models the adversary's decisions inspired by FP.
 \textcolor{black}{  We estimate the required probabilities with the empirical frequencies 
 of the opponent past plays 
 with $Q(a_i, b_j)$ updated according to the stateless version of \eqref{eq:lr}.
We refer to it as
FP$Q$-learning,}\footnote{\textcolor{black}{ Observe that only the DM uses such procedure to
model her opponent. In the standard FP algorithm \citep{brown1951iterative} all players 
are assumed to adopt such scheme which, under certain conditions,  converges to a
Nash equilibria; see, e.g., \citet{menache2011network}.}}
\textcolor{black}{ re-framing this in a Bayesian manner}
\textcolor{black}{ to 
favor its convergence should we have 
prior information about the adversary's decision making. 
Let $p_j = p_A(b_j)$ be the probability with which
the opponent chooses action $b_j$. 
If we adopt a Dirichlet prior
$(p_1 , \ldots, p_n) \sim \mathcal{D}(\alpha_1,\ldots,\alpha_n)$,
where $n$ is the number of actions available to the opponent, 
the posterior will be  
$\mathcal{D}(\alpha_1 + h_1,\ldots,\alpha_n + h_n)$, with  $h_i$ 
being the count for action $b_i$, $i=1,...,n$. 
If we denote the posterior density as $f(p|h)$, the DM would choose  action $a_i$ maximizing her expected reward,
adopting the form
\begin{eqnarray*}
& \psi(a_i) & = \int_{p \in \Delta^{n-1}} \left[\sum_{b_j \in \mathcal{B}}
Q(a_i, b_j) p_j\right] f(p|h) dp \\
&=& \sum_{b_j \in \mathcal{B}} Q(a_i, b_j) \mathbb{E}_{p|h}[p_j]
\propto  \sum_{b_j \in \mathcal{B}} Q(a_i, b_j) (\alpha_j + h_j).
\end{eqnarray*}
\textcolor{black}{where $\Delta^{n-1}$ denotes the standard $(n-1)$ simplex.}}
\vspace{-0.1in}

\textcolor{black}{
\paragraph{Including states} Generalizing this to account for states is 
conceptually straightforward. 
The $Q$-function adopts the form $Q(s, a_i, b_j)$. The DM needs to assess the probabilities $p_A(b_j | s)$,
  being natural to expect that her opponent behaves differently depending
on the state.
However, as $\mathcal{S}$ may be huge, even continuous, keeping track of $p_A(b_j|s)$ 
incurs in potentially prohibitive memory costs.}
\textcolor{black}{ 
Bayes rule $p_{A}(b_j| s) \,\, \propto \,\, p(s| b_j)p(b_j)$ 
 mitigates this problem, since the supported DM } will choose her action at state $s$ by maximizing
\[ \psi_s(a_i) = \sum_{b_j \in \mathcal{B}} Q(s, a_i, b_j) p_{A}(b_j|s)\,
\propto 
\sum_{b_j \in \mathcal{B}} Q(s, a_i, b_j) p_{A}(s | b_j) p (b_j ).  \]
\textcolor{black}{   Inspired by \citet{tang2017exploration}, 
  we propose keeping 
  track of $n$ bloom filters, one per 
 $p(s|b_j)$ distribution, to tractably predict the opponent's intentions
in the TMDP setting, with each filter maintaining
a count of the number of times that the agent visits each state.
 This is transparently integrated
within the Bayesian paradigm, as we only need to store an additional array with the Dirichlet prior parameters $\alpha_i$, $i=1,\ldots, n${\color{black},} for the $p(b_j)$ part.} 

\textcolor{black}{ As a final remark, we could straightforwardly extend the above scheme if we assume that the opponent has memory of the previous stage actions. However,
to mitigate memory requirements we use the concept of
 Markov chain mixtures \citep{raftery1985model}. Thus, we avoid \textcolor{black}{ the exponential growth in 
the number of required parameters and control
model complexity linearly}.
For example,
in case the opponent belief model
is $p_{A}(b_t | a_{t-1}, b_{t-1}, s_t)$, so that the adversary recalls 
 the previous actions $a_{t-1}$ and $b_{t-1}$, we \textcolor{black}{ would 
 factor }it through a mixture 
\[
p_{A}(b_t | a_{t-1}, b_{t-1}, s_t) = w_1 p_{A}(b_t | a_{t-1})  + w_2 p_{A}(b_t | b_{t-1})
 + w_3 p_{A}(b_t | s_t),
 \] 
 with $\sum_i w_i = 1, w_i \geq 0, i=1,2,3$.}
\textcolor{black}{ 
\subsection{Level-$k$ opponents}\label{sec:k}
When the opponent is 
  more 
strategic, he may model our supported DM as a level-0 thinker, thus making
him a level-1 thinker in the \citet{stahl1994experimental} sense. This chain can 
go up as needed, so we will
have to deal with modeling the opponent as a level-$k$ thinker, with $k$
bounded by \textcolor{black}{ the DM's computational or cognitive resources}.
For this, a hierarchy of TMDPs is introduced{\color{black},} with  
$\emph{TMDP}_{i}^k$ referring
to the TMDP that agent $i$, the DM or the 
adversary, needs to optimize,
while considering its rival as a level-$(k-1)$ thinker.
Therefore,  }
\textcolor{black}{
\begin{itemize}
\item If the supported DM is a level-1 thinker, she optimizes  $ \emph{TMDP}_{A}^1 $,
modeling $B$ as a level-0 thinker
(e.g., as in Section \ref{sec:non}).
\item If she is a level-2 thinker, the DM optimizes 
$ \emph{TMDP}_{A}^2 $ modeling $B$ as a level-1 thinker:
this ``modeled" $B$ optimizes $ \emph{TMDP}_{B}^1 $, \textcolor{black}{  who, to do so,
 models the DM as level-0.}
\item In general, we shall have a chain of TMDPs
$$ \emph{TMDP}_{A}^k \rightarrow \emph{TMDP}_{B}^{k-1}
\rightarrow \emph{TMDP}_{A}^{k-2}  \rightarrow \cdots $$
\end{itemize}
Exploiting the fact that TMDPs correspond to repeated interaction settings
(and, by assumption, both agents observe all past
decisions and rewards), each agent may  estimate their
counterpart's $Q$-function:  
   if the DM is optimizing $\emph{TMDP}_A^k$, she will maintain her own 
$Q$-function for this, call it $Q_k$, {\color{black} as well as } an estimate
$\hat{Q}_{k-1}$ of her opponent's $Q$-function. This estimate may be
computed optimizing $\emph{TMDP}_B^{k-1}$, and so on, until $k=1$.
The top level DM's policy is 
\[
\argmax_{a_{i_k}} Q_k(s, a_{i_k}, b_{j_{k-1}}),
\]
where $b_{j_{k-1}}$ is given by 
$
\argmax_{b_{j_{k-1}}} \hat{Q}_{k-1}(s, a_{i_{k-2}}, b_{j_{k-1}}),
$ and so on, until the induction basis (level-1) is reached 
in which the opponent may be modeled using the FP$Q$-learning approach 
from Section \ref{sec:non}.}
\textcolor{black}{ Figure
\ref{fig:lev2_scheme} provides a schematic view of the case in which
$k=2$, with the need to account for $Q_2$,  the DM's $Q$-function,  and $\hat{Q}_1$,
that of her
opponent (who will be level-1). }
\begin{figure}[!ht]
\centering
\stackinset{c}{.5in}{t}{.73in}{%
  \fboxrule=0pt\relax\framebox[2in][t]{%
  }}{\fboxrule=.75pt%
  \fbox{\stackunder{Level-2%
   \hspace*{\fill} (DM, denoted $A$) }%
    {
    $Q_2$,
    $p_A(b | s) \leftsquigarrow$
    \fbox{\stackunder{Level-1 \hspace*{\fill} (Adv., denoted $B$) }%
      {
      $\hat{Q}_1$,
    $p_B(a | s) \leftsquigarrow$
      \fbox{\stackunder{Level-0 \hspace*{\fill} (DM) }%
        {}
        }}}}}
}
\caption{Level-$k$ thinking scheme, with $k=2$}\label{fig:lev2_scheme}
\end{figure}

\noindent
\textcolor{black}{ Algorithm \ref{alg:l2ur} specifies the approach 
for a level-2 DM in detail, with immediate extension to higher level DMs}{\color{black}.}

\textcolor{black}{
\begin{algorithm*}[!ht]
\small
\begin{algorithmic}
\Require $Q_2$, $\hat{Q}_1$, $\alpha_2, \alpha_1$ (DM and opponent $Q$-functions, learning rates).
\State Observe  transition $(s, a, b, r_A, r_B, s')$ 
\State $\hat{Q}_1(s,b,a) := (1 - \alpha_1)\hat{Q}_1(s,b,a)  + \alpha_1 (r_B + \gamma \max_{b'} \mathbb{E}_{p_B(a'|s')} \left[ \hat{Q}_1(s',b', a') \right] )$ 
\State Compute $B$'s estimated $\epsilon$-greedy policy $p_A(b|s')$ from $\hat{Q}_1(s,b,a)$
\State $Q_2(s,a,b) := (1 - \alpha_2)Q_2(s,a,b) + \alpha_2 (r_A + \gamma \max_{a'} \mathbb{E}_{p_A(b'|s')} \left[ Q_2(s',a',b') \right]) $ 
\end{algorithmic}
\caption{Level-2 thinking update rule}
\label{alg:l2ur}
\end{algorithm*}
\noindent 
}

\textcolor{black}{
 Note that, in the previous hierarchy, decisions are obtained 
in a greedy manner, by maximizing the lower level
$\hat{Q}$ estimate. We may gain insight in a Bayesian fashion by adding uncertainty to the policy at each level. For instance, 
when considering $\epsilon$-greedy policies,
we could impose distributions $p_k(\epsilon)$ at each level $k$ of the hierarchy,
with the mean of $p_k(\epsilon)$ being an increasing function with respect to 
the level $k$ to account for the fact that uncertainty is higher
at the higher thinking levels.} 

\textcolor{black}{ 
\subsection{Combining Opponent Models}\label{sec:com}
 In most situations the DM will not actually know which type of particular opponent she is actually facing.
To deal with this, she may place a prior $p(M_i)$ denoting her beliefs that her opponent is using a model $M_i$, for $i = 1, \ldots, m$, the range of models that might describe her adversary's behavior,
with $\sum_{i=0}^m p(M_i) = 1$ and $p(M_i) > 0$,
$i=1,...,m$. 
As an example, she might place a Dirichlet prior on the levels of the $k$-level hierarchy. Then,
 after having observed her opponent's action at each iteration, she may
update her belief $p(M_i)$ by increasing the count $n_i$ of model $M_i$ which
caused that action, as in the standard Dirichlet-Categorical Bayesian
update rule, see Algorithm \ref{alg:update_averaging}.} 
\begin{algorithm*}[!ht]
\small
\begin{algorithmic}
\Require $p(M | H)\, \propto\, (n_1, n_2, \ldots, n_m)$ (counts for each model), with 
$H$ the sequence $(b_0, b_1, \ldots, b_{t-1})$ of past opponent actions.
\State At iteration $t$, observe  transition $(s_{t}, a_t, b_t, r_{A,t}, r_{B,t}, s_{t+1})$. 
\State For each opponent model $M_i$, set  $b^i$ to be the predicted action by model $M_i$.
\State If $b^i = b_t$, then update posterior
$$
p(M | (H || b_t) ) \, \propto\,  (n_1, \ldots, n_i + 1, \ldots, n_m) 
$$
\end{algorithmic}
\caption{Opponent average updating}
\label{alg:update_averaging}
\end{algorithm*}
\noindent \textcolor{black}{ This is possible since 
the DM maintains an estimate $p_{M_i}(b|s)$ of the opponent's policy for each
opponent model $M_i$. Should none of these have 
predicted the observed $b_t$, then we \textcolor{black}{ either do not perform an 
update, as in Algorithm \ref{alg:update_averaging} and done
in our experiments, or increase the count for all models.}}

\textcolor{black}{ 
This model averaging scheme subsumes the cognitive hierarchies framework \citep{camerer2004cognitive},   although the distribution is placed over the different hierarchy levels. Our scheme is more flexible, though,  since further kinds of opponents could be taken into account, 
say a minimax agent as 
in \citet{rios1}}.

\subsection{Approximate Q-learning with function approximation}\label{sec:approx}

\textcolor{black}{Tabular $Q$-learning (Algorithm \ref{alg:l2ur}) does not scale with the state 
or action spaces sizes.
\textcolor{black}{To address this issue, we extend the framework to settings in which $Q$-functions are represented using a function estimator, typically} }
\textcolor{black}{ a linear regression model or
a deep $Q$-network \citep{mnih2015human}. 
Algorithm \ref{alg:l2urdeep} reflects the corresponding details, where $\phi_A$ and $\phi_B$ designate the parameters of the corresponding functions approximating the $Q$-values. It resembles } 
Algorithm \ref{alg:l2ur}, \textcolor{black}{ but uses gradient descent for 
  updating purposes, as the parameters of the $Q$-networks are now continuous. Observe that
  this framework would be compatible with continuous action spaces.}

\begin{algorithm*}[!ht]
\small
\begin{algorithmic}
\Require  $Q_{\phi_A}$,   $\hat{Q}_{\phi_B}$, $\alpha_1, \alpha_2$ (DM's $Q$-function,
 estimate of opponent's $Q$-function, learning rates).
\State Observe  transition $(s, a, b, r_A, r_B, s')$.
\State $\phi_B := \phi_B - \alpha_1 \frac{\partial \hat{Q}_{\phi_B}}{\partial \phi_B}(s, b,a)\left[ \hat{Q}_{\phi_B}(s,b,a) - (r_B + \gamma \max_{b'}\mathbb{E}_{p_B(a'|s')} \hat{Q}_{\phi_B} (s', b',a') ) \right]  $
\State Compute $B$'s estimated $\epsilon$-greedy (or softmax) policy $p_A(b|s')$ from $\hat{Q}_{\phi_B}(s,b,a)$.
\State $\phi_A := \phi_A - \alpha_2 \frac{\partial Q_{\phi_A}}{\partial \phi_A}(s, a,b) \left[ Q_{\phi_A}(s,a,b) - (r_A + \gamma \max_{a'}\mathbb{E}_{p_A(b'|s')} Q_{\phi_A} (s', a',b') ) \right]  $ 
\end{algorithmic}
\caption{Level-2 thinking update rule using function estimators.}
\label{alg:l2urdeep}
\end{algorithm*}
\vspace{-0.1in}

\textcolor{black}{
\subsection{Facing multiple opponents}\label{sec:mul}
  \textcolor{black}{ As announced, TMDPs may be extended to cases in which the DM faces more than one adversary. If so,} the DM would have uncertainty about all of her opponents and she would need to average her $Q$-function over their
likely actions. Let $p_A(b^1, \dots, b^M \vert s)$ represent the DM's beliefs about her $M$ adversaries' actions. The extension of the TMDP framework to multiple adversaries will require to account for all possible opponents in the DM's $Q$ function which adopts the form $Q(s,a,b^1,\dots,b^M)$. Finally, the DM would average this $Q$ over $b^1, \dots, b^M$ in \eqref{eq:lr}, proceeding as in \eqref{eq:lr2}.}

\textcolor{black}{ When the DM faces non-strategic opponents, she could learn $p_A(b^1, \dots,$ $ b^M \vert s)$ in a Bayesian way, \textcolor{black}{ as in Section \ref{sec:non}.} This would entail placing a Dirichlet prior on the $n^M$ dimensional vector of joint actions of all adversaries. However, keeping track of those probabilities may be unfeasible as the dimension scales exponentially with the number of opponents. When  
  $A$ believes her adversaries act conditionally independent given the
state, the analysis is simpler, as, in this case, 
$p_A(b^1, \dots, b^M \vert s) = p_A(b^1 \vert s) \dots p_A(b^M \vert s)$. Then, we could learn each $p_A(b^i \vert s)$ for $i=1, \dots, M$ separately as above,     then combining the forecasts  
multiplicatively.}

\textcolor{black}{
\subsection{Computational complexity}\label{sec:cc}
Let us finish with a brief discussion on computational complexity  
\textcolor{black}{ drawing on the Section 3.2 approach: a }
level-$k$ $Q$-learner has to estimate the $Q$ function
of a level-$(k-1)$ $Q$-learner, \textcolor{black}{ which has to learn the 
$Q$ function of a level-$(k-2)$ $Q$ learner,} and so on. Assuming that the original $Q$-learning 
update rule has time complexity $\mathcal{O}(T(|\mathcal{A}|))$, with $T$ 
  a factor that depends on the \textcolor{black}{ number of DM's actions,  
Algorithm \ref{alg:l2ur} update rule} has time complexity $\mathcal{O}(kT(\max \lbrace |\mathcal{A}|,  |\mathcal{B}|\rbrace))$, i.e.,
it is linear in the level of the hierarchy.} 

\textcolor{black}{
Regarding space complexity, the overhead is also linear in such level $k$ since the DM only needs to store $k$ $Q$-functions, leading to a memory complexity $\mathcal{O}(kM(|\mathcal{S}|,|\mathcal{A}| \cdot |\mathcal{B}|))${\color{black},} with $M(|\mathcal{S}|,|\mathcal{A}| \cdot |\mathcal{B}|)$ accounting for the memory needed to store the $Q$-function in tabular form. This quantity depends on the number of states and pairs of actions for the DM and her opponent.
}

 Appendix B provides further computational verification. Timings in Table~\ref{tab:times_combined} corroborate the linear-in-$k$ bound: per-experiment runtime grows roughly linearly across levels $0$--$3$ in the games in Section~\ref{sec:statv}, Level2Q incurs only a $\sim$$1.6\times$ overhead over IndQ in the Blotto experiment, and RegressionLevel2Q is about twice as costly as RegressionIndQ in the gridworld of Section~\ref{sec:cg}.  The multi-opponent extension scales provided the conditional-independence factorization of Section~\ref{sec:mul} is exploited.
\vspace{-0.1in}

\section{Experiments}
\textcolor{black}{Let us illustrate the main modeling and computational concepts of the proposed TMDP reasoning framework through four sets of experiments.  First, 
an adversarial security environment proposed by  \citet{leike2017ai}
  used to illustrate robustness issues. 
 Next, a Blotto game security resource allocation problem to show how to handle multiple opponents. Third, a 
gridworld game to showcase the compatibility 
with $Q$-values estimated with parametric functions, as in deep RL.
 Finally, computational comparisons with other approaches are highlighted.}

\begin{figure}
    \centering
    \vspace{-\baselineskip}
    \includegraphics[scale=0.4]{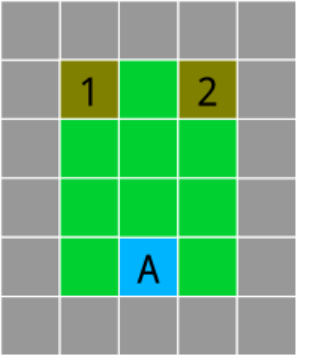}
    \caption{\emph{Friend or foe} environment from AI Safety Gridworlds {\color{black}\citep{leike2017ai}}.
    Blue cell represents DM's initial state, gray cells represent room walls.}
    \label{fig:friendorfoe}
\end{figure}

\textcolor{black}{
\subsection{Advantages of modeling adversaries}\label{s:sg}
 Let us demonstrate first that by
explicitly modeling the behavior of adversaries, our
 framework improve upon standard $Q$-learning methods. 
For this, we consider a suite of RL safety benchmarks
  in \citet{leike2017ai}, focusing  
on the safety \emph{friend or foe} environment:
the supported DM
needs to travel through a room and choose 
between two identical boxes, respectively hiding a positive (+50)
and a negative (-50) reward, under control by
an adaptive opponent. Figure \ref{fig:friendorfoe} shows the initial set up. 
Cells 1 and 2 depict the adversary's targets, who decides which 
one will 
hide the positive reward.}

\textcolor{black}{ This may be viewed as a spatial Stackelberg game in which
the adversary plans to attack one of two targets. The defender
  obtains a positive reward if she travels to the chosen target. Otherwise,
she misses the attacker and incurs in a loss.
As \citet{leike2017ai} show, a \emph{deep $Q$-network}
fails to
achieve optimal results since the reward process is controlled by the adversary.}

\textcolor{black}{
\subsubsection{Stateless Variant}\label{sec:statv}
Consider first a simplified setting with a single state and
two actions. As in  {\color{black}\citet{leike2017ai}, an } adaptive
opponent estimates the DM's actions using an exponential smoother.
Let $\bm{p} = (p_1, p_2)$ be the probabilities
with which the DM respectively chooses targets 1 or 2
as estimated by $B$. 
Initial estimates 
are
$\bm{p} = (0.5, 0.5)$; after each iteration, the update is 
$
\bm{p} := \beta \bm{p} + (1 - \beta ) \bm{a},
$
with $0 < \beta < 1$ 
a learning rate   unknown by the DM,
and $\bm{a} \in \lbrace (1, 0), (0, 1) \rbrace$ is a one-hot encoded
vector indicating whether the DM  chose target 1 or 2. 
Assume an adversary which places the positive reward in target
$t = \argmin_i (\bm{p})_i$.}

\textcolor{black}{
Since the DM has to deal with a strategic adversary, 
consider a variant of FP$Q$-learning (Section 3.1) giving
additional relevance to more recent actions.
Algorithm \ref{alg:duwff} provides a modified update 
 using the fact that the Dirichlet distribution is conjugate of the Categorical one. Instead of weighting all observations equally, we essentially account for just the last $\frac{1}{1 - \lambda}$ 
 opponent actions, where $0< \lambda < 1 $  is viewed as a forget factor.
 \begin{algorithm}
 \small
\begin{algorithmic}
\State Initialize pseudocounts $ \bm{\alpha^0} = (\alpha^0_1, \ldots, \alpha^0_n)$
\For{$t = 1, \ldots, T $}
\State $\bm{\alpha^t} = \lambda \bm{\alpha^{t-1}}$ \Comment Reweight with factor $0 < \lambda < 1$
\State Observe opponent action $b^t_i, i \in \lbrace b_1, \ldots, b_n \rbrace$
\State $\alpha^t_i = \alpha^{t-1}_i + 1$ \Comment Update posterior
\State $\alpha^t_{-i} = \alpha^{t-1}_{-i}$
\EndFor
\end{algorithmic}
\caption{Dirichlet updating with forget factor}
\label{alg:duwff}
\end{algorithm}
}

\textcolor{black}{
For a level-2 DM, as we do not know 
the actual adversary rewards (a level-1 learner), 
 we  model it as in a zero-sum scenario  ($r_B = -r_A$){\color{black},} making this case similar to the Matching Pennies game. The discount factor is $\gamma = 0.8$. There are $5000$ episodes. The initial exploration parameter
 is $\epsilon = 0.1$; the learning rate, $\alpha = 0.1$. The 
 forget factor is $\lambda = 0.8$.
\begin{figure}[hbt]
\centering
\begin{minipage}{0.48\textwidth}
    \centering
    \includegraphics[scale=0.32]{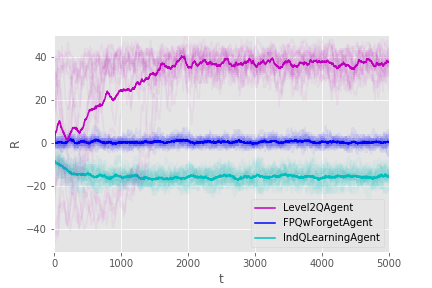}
    \caption{Rewards for DM against adversarial opponent}
    \label{fig:4C_adv}
\end{minipage}\hfill
\begin{minipage}{0.48\textwidth}
    \centering
    \includegraphics[scale=0.32]{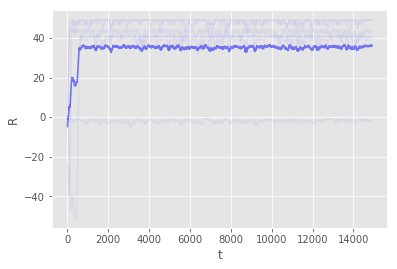}
    \caption{Rewards for DM against adversarial opponent, using softmax policy.}
    \label{fig:softmax}
\end{minipage}
\end{figure}
  Figure \ref{fig:4C_adv} displays results for three defenders:
an opponent unaware $Q$-learner (green); a level-1 DM with forget 
  factor 
using Algorithm \ref{alg:duwff} (blue); and, finally, a
level-2 agent (purple). The first defender 
 is clearly exploited by the adversary 
achieving suboptimal rewards (close to -20).
In contrast, the level-1 DM with forget {\color{black}factor } 
learns a stationary optimal policy (reward 0). Finally, \textcolor{black}{ the
level-2 agent 
exploits 
the adaptive adversary achieving positive rewards
(around 40, close to the upper bound of 50 due to the value of the positive reward)}.} 
\textcolor{black}{ 
Note that the actual adversary behaves differently from how the DM models him,
as he is not a level-1 $Q$-learner. Even so, modeling him as 
{\color{black}such gives} the DM sufficient advantage in this case.
  Such robustness against opponent misspecification emerges as an
advantage of our framework.}

\begin{figure}%
 \centering
 \subfigure[Rewards $+1$ and $0$ for the adversary]{%
 \label{fig:4C_binary_adv}%
 \includegraphics[scale=0.28]{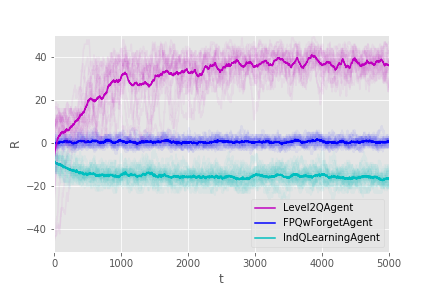}}%
 \subfigure[Rewards $+1$ and $-1$ for the adversary]{%
 \label{fig:4C_binary_1m1}%
 \includegraphics[scale=0.28]{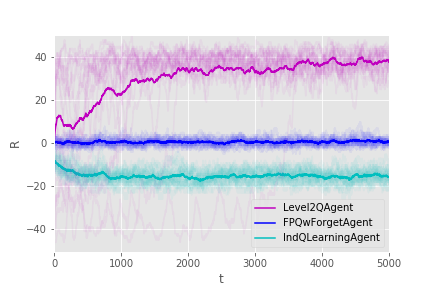}}%
 \caption{Rewards against same adversary (exp. smoother){\color{black},} using different reward scaling.}\label{fig:4C_rs}
 \end{figure}

\textcolor{black}{  Next we replace the $\epsilon$-greedy
policy by a softmax one: actions at state $s$ are taken with probabilities 
proportional to the $Q(s, a)$ values. Figure \ref{fig:softmax}
provides several simulation runs of a level-2 $Q$-learner versus the adversary: 
changing the policy} sampling scheme does not worsen
the DM with respect to the $\epsilon$-greedy alternative.
  \textcolor{black}{Lastly, we \textcolor{black}{ checked with other models } for the opponent's rewards $r_B$. \textcolor{black}{  Instead of 
the initial zero-sum setting, 
we tried two additional reward scalings $r_B \in \lbrace -1, 1 \rbrace$ and $r_B \in \lbrace 0, 1 \rbrace $. Figure \ref{fig:4C_rs} displays the results, portraying
similar qualitative behavior than in Figure 4. 
}}

\subsubsection{Facing more powerful adversaries}
 \textcolor{black}{ We study now cases when the DM faces adversaries more powerful 
 than exponential smoothers}.

\begin{figure*}[!ht]%
\centering
\subfigure[L2$Q$ vs.\ L2$Q$]{%
  \label{fig:L2vsL2}%
  \includegraphics[width=0.31\textwidth]{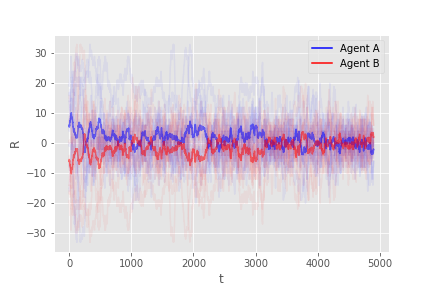}}%
  \subfigure[L3$Q$ vs.\ L2$Q$]{%
  \label{fig:L3vsL2}%
  \includegraphics[width=0.31\textwidth]{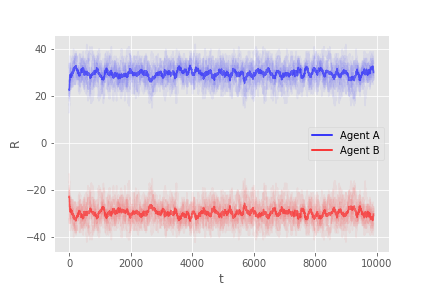}}%
  \subfigure[L3$Q$ vs.\ L1$Q$]{%
  \label{fig:L3vsL1}%
  \includegraphics[width=0.31\textwidth]{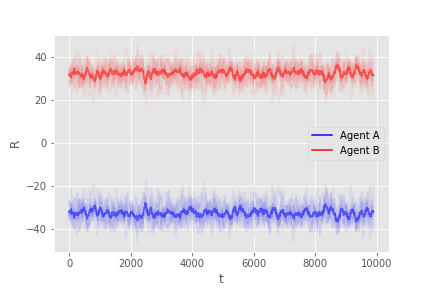}}\\
  \subfigure[L3$Q$ with opponent averaging vs.\ L1$Q$]{%
  \label{fig:L3DirvsL1}%
  \includegraphics[width=0.42\textwidth]{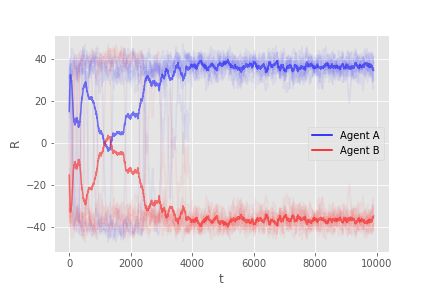}}%
  \subfigure[Estimate of $P_{L1Q}$]{%
  \label{fig:probas}%
  \includegraphics[width=0.42\textwidth]{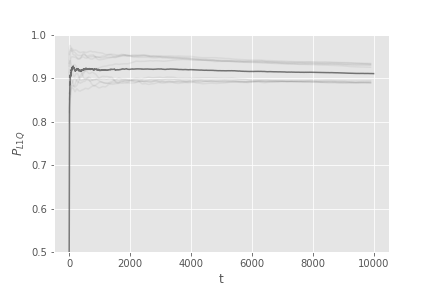}}%
  \caption{Rewards (a--d) and posterior belief (e) against an exponential-smoother adversary. (a--c): DM is blue, opponent red.}
\end{figure*}

First, let us parameterize our opponent as a level-2 $Q$-learner.
\textcolor{black}{ For simplicity we specify the rewards he receives as $r_B = -r_A$ as
in a zero-sum case, although the framework allows for  
 general-sum problems}. Figure \ref{fig:L2vsL2} depicts the rewards for both the DM (blue)
and her adversary (red). We have computed the frequency of choosing each of the actions, and both players select either action 
 with probability $0.5 \pm 0.002$ based on 10 different random seeds,
thus achieving a Nash equilibrium  (choosing between both
actions with equal probability), leading to an expected cumulative reward of 0. 

\textcolor{black}{ Increasing the DM sophistication to make her level-3 allows} her to exploit a level-2 adversary, Fig. \ref{fig:L3vsL2}. However, this DM fails to exploit a level-1 opponent,
Fig. \ref{fig:L3vsL1}. The explanation to this apparent paradox is that the DM is modeling her opponent as a more powerful agent
than he actually is: this inaccuracy leads to poorer performance.
However, this ``failure" suggests a potential solution to the problem
using type-based reasoning (Section \ref{sec:com}). Figure \ref{fig:L3DirvsL1} depicts the rewards of a DM that keeps track of both level-1 and level-2 opponent models and learns, in a Bayesian manner, which one she is  
actually facing. The DM keeps estimates of the probabilities $P_{L1Q}$ and $P_{L2Q}$ that her opponent is acting as if he was a level-1 or a level-2 $Q$-learner, respectively. Figure \ref{fig:probas} tracks the evolution of $P_{L1Q}$, \textcolor{black}{ which places most of the probability in the correct opponent type.} 
\textcolor{black}{
\subsubsection{Spatial Variant}
\textcolor{black}{Compare now an independent $Q$-learner and a level-$2$ $Q$-learner against the
  adaptive exponential smoother opponent in the spatial gridworld domain
 from Fig. \ref{fig:friendorfoe}. The DM has four actions, one for each direction in which 
  she is allowed to move one step. Target rewards 
are delayed until the DM arrives at one of the pertinent locations, 
obtaining $\pm 50$.} 
Each step is penalized with -1 for the DM. Episodes end at a maximum of 50 steps or when the agent arrives first at targets 1 or 2. 
The discount factor is $\gamma = 0.8$ and $15000$ episodes are 
considered. For the level-2 agent, we consider different exploration hyperparameters, $\epsilon_A$ and $\epsilon_B${\color{black},} for the DM's policy and her corresponding model for her opponent;
initially,  $\epsilon_A = \epsilon_B = 0.99$ with decaying rules $\epsilon_A := 0.995\epsilon_A$ and $\epsilon_B := 0.9\epsilon_B$ every $10$ episodes{\color{black},} and learning rates {\color{black}$\alpha_1 = \alpha_2 = 0.05$}. For the independent $Q$-learner, we set similar initial hyperparameters.}

\begin{figure*}[h]
\centering
\subfigure[Rewards for various DM models]{%
  \label{fig:4C_gridworld}%
  \includegraphics[height=1.3in]{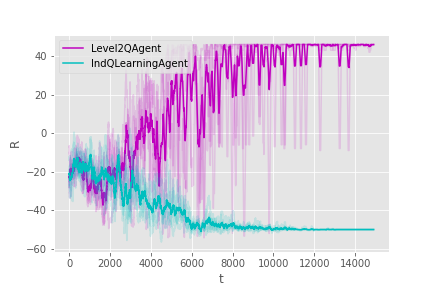}}%
  \subfigure[Rewards for a DM with opponent models for a L1 $Q$-learner and a L2 $Q$-learner (blue)]{%
  \label{fig:L3Dir_spatial}%
  \includegraphics[height=1.3in]{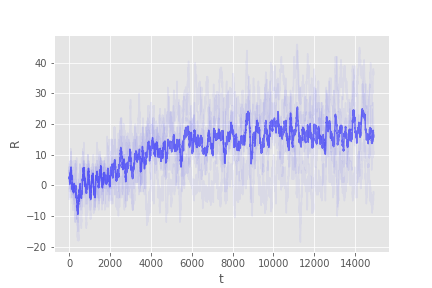}}%
  \caption{Rewards against exponential smoother opponent in the spatial environment. }
\end{figure*}

\textcolor{black}{
\textcolor{black}{ Figure \ref{fig:4C_gridworld} displays results}. Again, an independent $Q$-learner \textcolor{black}{(green}) is
exploited by the adversary, obtaining even more negative results  
than in 
Figure \ref{fig:4C_adv} due to the penalty at each step. In contrast,
a level-2 agent \textcolor{black}{(purple}) approximately estimates adversarial
behavior, modeling him as a level-1 agent, obtaining positive rewards. Figure \ref{fig:L3Dir_spatial} depicts the rewards of a DM that maintains 
opponent models for both level-1 and level-2 $Q$-learners. Although the adversary is of neither class, the DM achieves positive rewards,
suggesting that the framework is capable of generalizing between opponent behaviors not exactly reflected by the DM's opponent model.
This suggests that our level-$k$ thinking scheme is sufficiently robust to opponent misspecification.}

\textcolor{black}{Lastly, we perform experiments with different hyperparameter values,
  to further highlight the \textcolor{black}{framework robustness.}
  \textcolor{black}{ Table \ref{tab:rob} displays mean rewards and standard deviations for five different random seeds,
    over different hyperparameter choices within Algorithm \ref{alg:l2ur}. Except when the initial exploration rate $\epsilon_0$ is set to a high value} (0.5, which makes the DM achieve a positive mean reward), the other settings showcase that the framework (for the level-2 case) is robust to different learning rates.} 

\begingroup
\renewcommand{\arraystretch}{0.7}
\begin{table}[h]
\footnotesize
\caption{Hyperparameter robustness of Algorithm \ref{alg:l2ur} in the spatial gridworld.}\label{tab:rob}
\centering
\begin{tabular}{lllr}
\hline
$\alpha_1$ &  $\alpha_2$ & $\epsilon_0$ & Mean Reward \\
\hline
 0.005 & 0.01 & 0.5 & $15.46 \pm 47.21$  \\
 0.005 & 0.01 &0.1 & $40.77 \pm 27.48$  \\
 0.005 & 0.01 & 0.01 & $46.32 \pm 16.15$  \\\hline
 0.02 & 0.01 & 0.5 & $15.58 \pm 47.17$  \\
 0.02 & 0.01 & 0.1 & $43.05 \pm 23.65$  \\
 0.02 &0.01 & 0.01 & $47.81 \pm 10.83$  \\\hline
 0.05 & 0.1 & 0.5 & $15.30 \pm 47.27$  \\
 0.05 &0.1 & 0.1 & $42.82 \pm 24.08$  \\
 0.05 &0.1 & 0.01 & $48.34 \pm 8.10$  \\\hline
 0.2 & 0.1 & 0.5 & $15.97 \pm 47.03$  \\
 0.2 & 0.1 & 0.1 & $43.05 \pm 23.66$  \\
 0.2 & 0.1 & 0.01 & $48.51 \pm 6.96$  \\\hline
  0.25 & 0.5 & 0.5 & $15.95 \pm 47.04$  \\
  0.25 & 0.5 & 0.1 & $43.06 \pm 23.64$  \\
  0.25 & 0.5 & 0.01 & $48.41 \pm 7.68$  \\\hline
  1.0 & 0.5 & 0.5 & $15.19 \pm 47.31$  \\
  1.0 & 0.5 & 0.1 & $42.98 \pm 23.71$    \\
  1.0 & 0.5 & 0.01 & $48.53 \pm 6.82$  \\\hline
\end{tabular}
\end{table} 
\endgroup
\subsection{Facing multiple opponents}
\textcolor{black}{ Let us illustrate the multiple opponent concepts from Section \ref{sec:mul} introducing a novel suite of  security resource allocation experiments. 
 They are based on a modified Blotto game \citep{hart2008discrete}\textcolor{black}{, a security resource-allocation setting also addressed, e.g., via counterfactual regret minimization in  \citet{keith2021counterfactual}}. 
 A DM needs to distribute limited
resources over several positions susceptible of being attacked. Each of the attackers 
 chooses different positions
to deploy their attacks. Associated with each of the attacked positions there is a positive (resp. negative) reward with value 1 (-1). If the DM 
deploys more resources than the attackers in a given position, she wins the positive reward and the negative is equally split  between
the attackers choosing to attack such position. If the DM deploys less resources,
 she receives
  a negative reward and the positive one will be equally split between the corresponding attackers.
In case of  draw, no player receives any reward.} 

\begin{figure}[h]
\centering
\begin{minipage}{0.5\textwidth}
    \centering
    \includegraphics[scale=0.35]{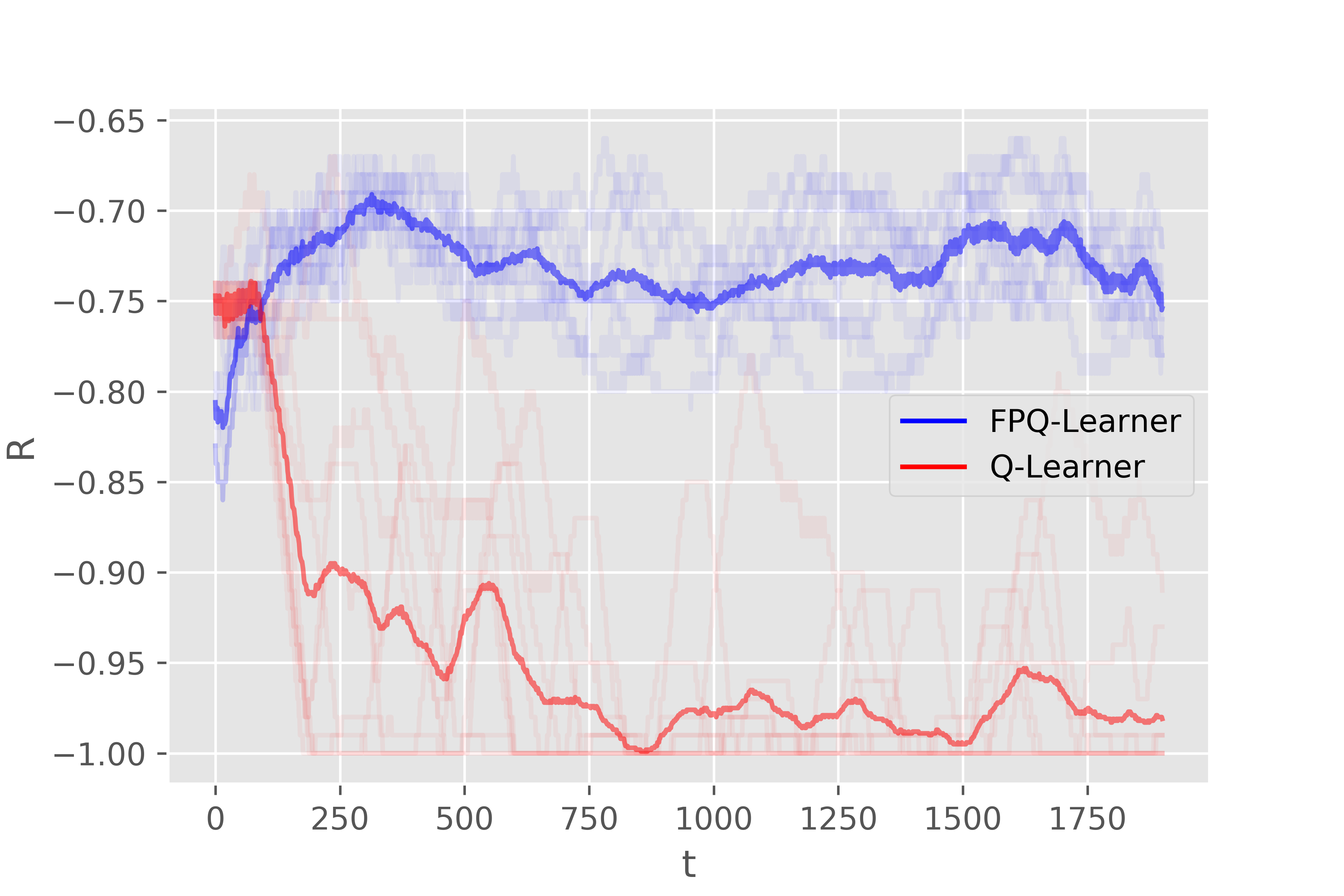}
    \caption{Rewards for the DM against the opponents}
    \label{fig:2expsmoothers}
\end{minipage}\hfill
\begin{minipage}{0.45\textwidth}
    \centering
    \includegraphics[scale=0.2]{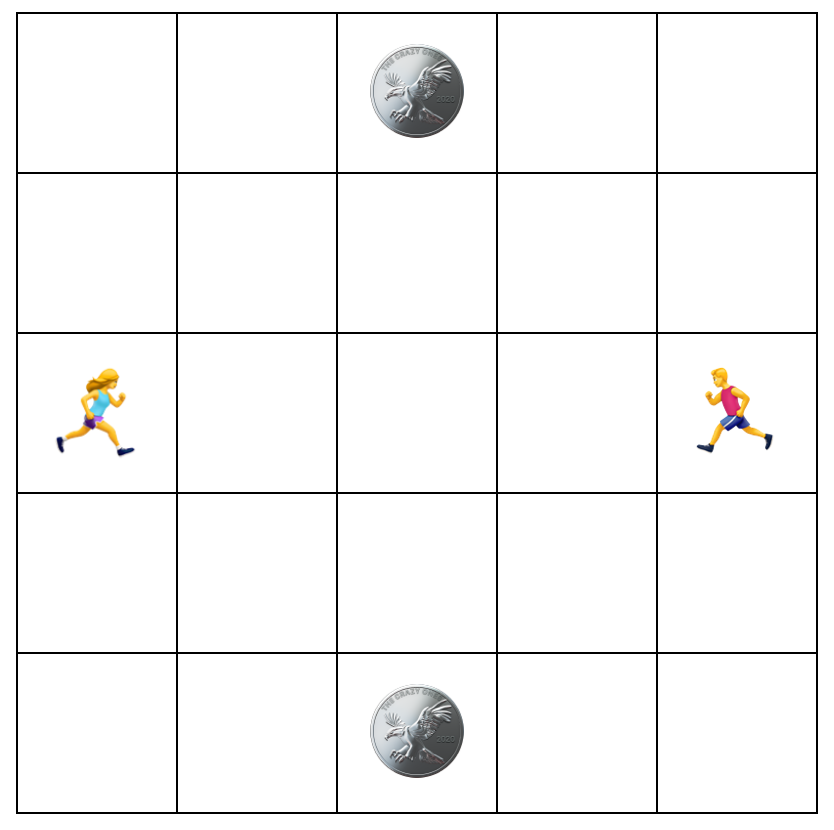}
    \caption{An initial state for the Coin Game}
    \label{fig:coingame}
\end{minipage}
\end{figure}

\textcolor{black}{ 
We compare the performance of an FP$Q$-learner versus a standard $Q$-learner, when facing two conditionally independent opponents{\color{black},} both using 
exponential smoothing to estimate the probability of the DM placing a resource at each position, and implementing the attack where
this probability is the smallest (obviously, both opponents perform exactly the same attacks).}
\textcolor{black}{
Consider
 defending three different positions. The DM needs to allocate two resources among such positions. For both the $Q$- and the FP$Q$-learner{\color{black},} the parameters will be $\gamma = 0.96$, $\epsilon = 0.1$, and $\alpha = 0.1$. 
 As Fig.\ \ref{fig:2expsmoothers} shows, FP$Q$-learning (blue) is able to learn the opponents strategies and thus is less exploitable than standard $Q$-learning (red).
This experiment suggests the suitability of our proposal to deal with multiple, independent adversaries, by extending the level-$k$ thinking scheme as 
in Section \ref{sec:mul}.}
%







\subsection{Experiments with function approximation}\label{sec:cg}

\textcolor{black}{
  We now run experiments to showcase 
 that the framework is compatible with variants in which $Q$-values are approximated with a parametric function, in particular a deep learning model, recall Section \ref{sec:approx}.}
\noindent 
\textcolor{black}{ The game in Figure \ref{fig:coingame}  
is played over an $N \times N$ grid ($N=5$ in the Figure). Each turn, both players (blue and red) move in either of the four directions (unless they try to cross the boundary of the grid), with the objective of capturing two different coins. The episode ends after 12 steps, so that each agent has ample time to pick both coins. For each coin an agent picks, he receives a reward of 1, unless the other agent arrives also at the same location,
in which case the perceived reward is 0.5. 
  Thus, this game can be cast as a coordination game.} 

\textcolor{black}{ The space of states is  modeled as an $N \times N \times 4$ array. Each of 
its four last slices denotes the position of both players and both coins as a one-hot encoded $N \times N$ matrix. A straightforward application of $Q$-learning would require a
 $Q(s, a)$ table with $(N \times N)^4 \times 4$ entries. Instead, we parameterize the $Q$-values to reduce the number of parameters by flattening the state $s$ 
 from a tensor in $\mathbb{R}
^{N \times N \times 4}$ to a vector in $\mathbb{R}
^{4N^2}$, and projecting it into $\mathbb{R}^4$ using linear regression to obtain one $Q$-value for each of the four possible actions. Given any state $s \in \mathbb{R}^{4N^2}$, we obtain the corresponding $Q$-value 
via $Q(s, a) = w_a^{\intercal}s ${\color{black},} with $w_a \in \mathbb{R}^{4N^2}$ being the weight vector for action $a$. Thus, we reduce the number of parameters to $N
^2\times 4 \times 4$. The extension to the $Q$-function estimator for the level-2 case is straightforward. Rather than projecting to $\mathbb{R}^4$, 
we do it to $\mathbb{R}^{4\times 4}$ to consider each pair of actions $(a, b)$, to approximate the corresponding $Q(s,a,b)$ function.} 

\begin{figure*}%
\centering
\subfigure[Rewards of two independent $Q$-learners against each other]{%
  \label{fig:coin1}%
  \includegraphics[height=1.3in]{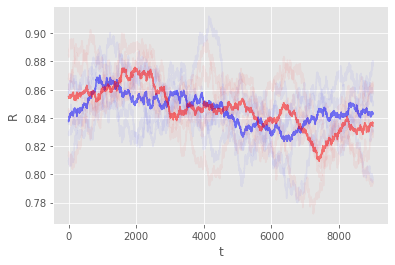}}%
  \subfigure[Rewards of a L2 $Q$-learner (blue) against an independent $Q$-learner (L0) (red)]{%
  \label{fig:coin2}%
  \hspace{0.2cm}
  \includegraphics[height=1.3in]{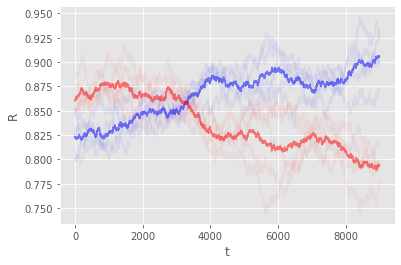}}%
  \caption{Rewards in the Coin Game using function estimators. }
\end{figure*}

\textcolor{black}{ 
Figures \ref{fig:coin1} and \ref{fig:coin2} depict the results of two games. Light color
  depicts the smoothed rewards over 10000 iterations with five random seeds.  Darker colors reflect the three averages of the $3 \times 5$ previous runs. Even in this approximate regime, a DM exhibiting higher rationality than its adversary
can behave advantageously. Thus, our opponent modeling framework is compatible with approximate $Q$-values, being capable of harnessing all the benefits from the approximate regime, such as a reduced parameter dimensionality 
or enhanced generalization capabilities.}

\textcolor{black}{ Finally, we also used a multi-layer NN to parameterize the $Q$-functions, with no significant improvement in the results. Therefore, we just report those attained with the   linear model.}

\subsection{\textcolor{black}{Comparisons against canonical and recent multi-agent methods}}\label{sec:wolf}\label{sec:loqa}

\textcolor{black}{Sections~\ref{s:sg}-\ref{sec:cg} benchmarked our scheme against an independent (opponent-unaware) $Q$-learner, which is the natural ablation of our method: switching off the opponent-modeling component recovers exactly a standard $Q$-learner, so the comparison isolates the contribution of opponent modeling. To complement this ablation,} we report head-to-head experiments \textcolor{black}{in an iterated Chicken (IC) game, whose payoff bimatrix is recalled in Table~\ref{tab:payoffC2}: the actions $C$ (cooperate) and $D$ (defect) yield two pure Nash equilibria, $(C,D)$ and $(D,C)$, with mutual defection $(D,D)$ giving the worst outcome $(-4,-4)$. We face the DM against two external opponents: WoLF-PHC \citep{bowling2001rational}, a canonical multi-agent learner, and LOQA \citep{aghajohari2024loqa}, a state-of-the-art representative of the \emph{opponent-shaping} family that has succeeded LOLA.}

\begin{figure*}[!ht]%
\centering
\subfigure[FP$Q$ (blue) vs.\ WoLF-PHC (red)]{%
  \label{fig:L1vsWoLF_C}%
  \includegraphics[width=0.31\textwidth]{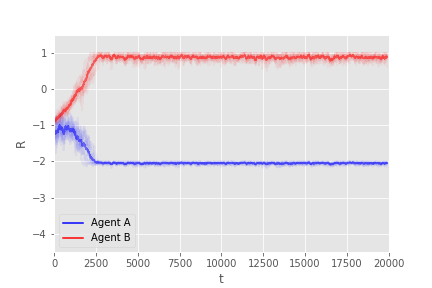}}%
  \subfigure[L2$Q$ (blue) vs.\ WoLF-PHC (red)]{%
  \label{fig:L2vsWoLF_C}%
  \includegraphics[width=0.31\textwidth]{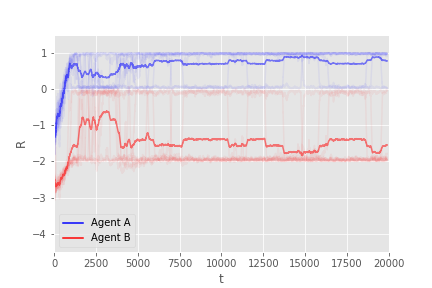}}\\
  \subfigure[FP$Q$ (blue) vs.\ LOQA (red)]{%
  \label{fig:L1vsLOQA_C}%
  \includegraphics[width=0.31\textwidth]{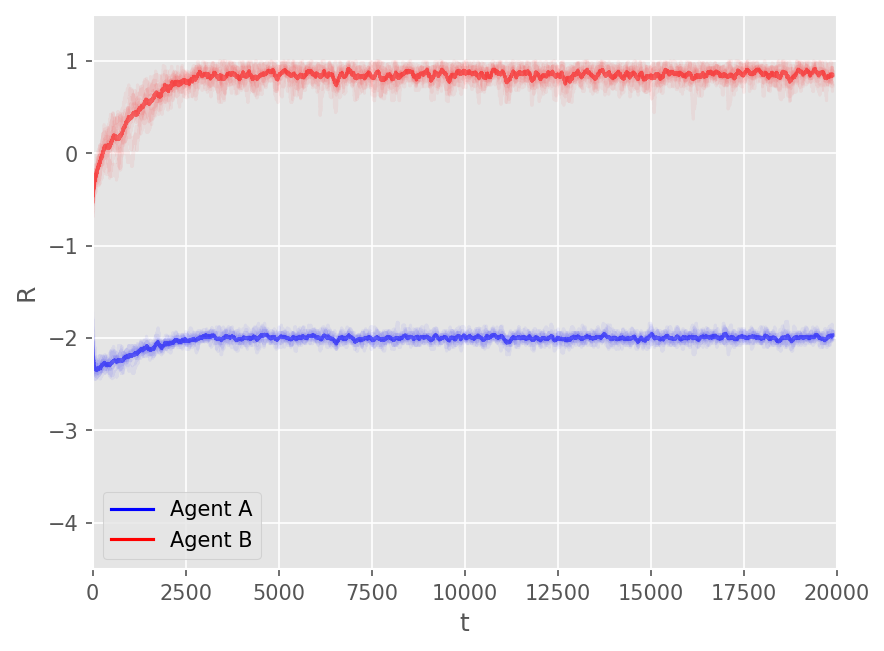}}%
  \subfigure[L2$Q$ (blue) vs.\ LOQA (red)]{%
  \label{fig:L2vsLOQA_C}%
  \includegraphics[width=0.31\textwidth]{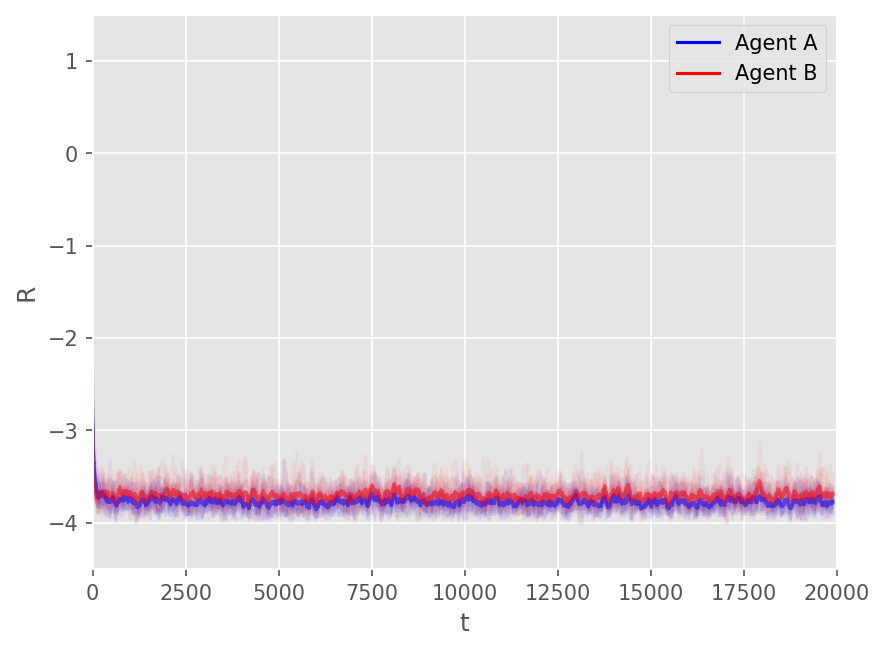}}%
  \subfigure[L3$Q$ with opp.\ avg.\ (blue) vs.\ LOQA (red)]{%
  \label{fig:L3DirvsLOQA_C}%
  \includegraphics[width=0.31\textwidth]{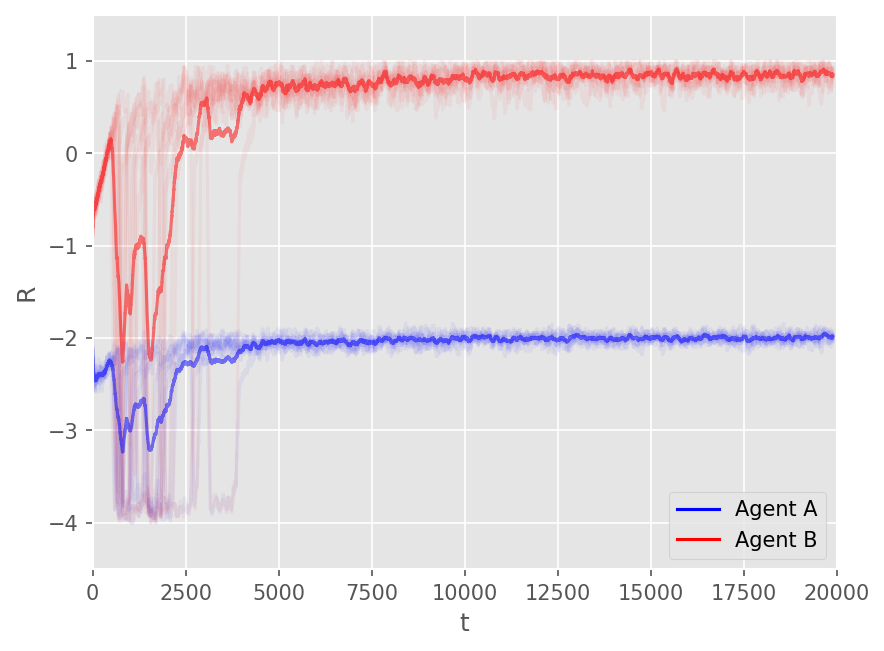}}%
  \caption{\textcolor{black}{Rewards in the iterated Chicken game against WoLF-PHC (top row) and LOQA (bottom row) adversaries.}}
  \label{fig:loqa}
\end{figure*}

\begin{table}[h!]
\begin{center}
\begin{tabular}{c|c|c}
\hline
 & C & D \\
\hline
C & (0, 0) & (-2, 1) \\
\hline
D & (1, -2) & (-4, -4)  \\
\hline
\end{tabular}
\end{center}
\caption{\textcolor{black}{Payoff bimatrix of the Chicken game used in this subsection. Rows correspond to  DM's actions and columns to the opponent's. Each cell reports the (DM, opponent) rewards. Action 
 $C$ denotes \emph{cooperate}; action $D$, \emph{defect}. }}
\label{tab:payoffC2}
\vspace{-2ex}
\end{table}
\textcolor{black}{As a level-1 thinker (FP$Q$-learner) the DM is exploited by WoLF-PHC, who steers the joint dynamics toward its preferred $(C,D)$ (Figure~\ref{fig:L1vsWoLF_C}). Modeling the DM as level-2 reverses the outcome: she anticipates the opponent's adaptation and converges to her preferred $(D,C)$, dominating WoLF-PHC (Figure~\ref{fig:L2vsWoLF_C}).}

\textcolor{black}{LOQA, in turn, models its opponent as a $Q$-learner that adopts a softmax policy over its action values, estimates those values from the opponent's observed actions and rewards, and ascends the gradient of its expected return \emph{through} the opponent's induced policy, thereby shaping the opponent's learning to its own benefit. Unlike the other recent methods later discussed in Section~\ref{sec:discussion}, its informational requirements coincide exactly with those of TMDPs, enabling a clean comparison: we employ a tabular instantiation with Monte Carlo estimators are replaced by exact expectations.\footnote{\textcolor{black}{The adversary maintains running-mean estimates $\hat{r}_A(a,b)$, $\hat{r}_B(a,b)$ of both reward functions; models the DM as playing softmax over $\hat{Q}_A(a) = \sum_b \pi_{\theta}(b)\, \hat{r}_A(a,b)$, exact up to an action-independent shift for a stateless $Q$-learner at its fixed point; and ascends the exact gradient of $V_B(\theta) = \sum_a \hat{\pi}_A(a;\theta) \sum_b \pi_\theta(b)\, \hat{r}_B(a,b)$, which includes LOQA's opponent-shaping path through $\hat{\pi}_A$. Results are robust to its learning rate, softmax temperature, and exploration rate.}} Under the same protocol, LOQA does exactly what it was designed for: against level-0 and level-1 DMs it commits to defection, steering our DM's $Q$-values until she concedes to its preferred $(C,D)$ (Figure~\ref{fig:L1vsLOQA_C}). A level-2 DM cannot be shaped: she too expects the opponent to concede, but LOQA never does, and both shapers lock into mutual defection (Figure~\ref{fig:L2vsLOQA_C}), an instance of a shaper-versus-shaper conflict \citep{lu2022model}. Finally, the type-based DM with opponent averaging (Section~\ref{sec:com}) identifies a committed defector and concedes, securing the maximal $-2.0$ (Figure~\ref{fig:L3DirvsLOQA_C}). POLA \citep{zhao2022proximal}, the proximal successor to LOLA run as an adversary estimating the DM's policy from play, yields the same pattern (trajectories in the companion code), so the advantage extends to the modern shaping family as a whole. Taken together, no fixed cognitive level dominates: the level-2 DM prevails over the adaptive WoLF-PHC yet clashes with the shaping LOQA, and the practical advantage lies in Bayesian, type-based selection among opponent models.}

\begin{figure*}[!ht]%
\centering
\includegraphics[width=0.99\textwidth]{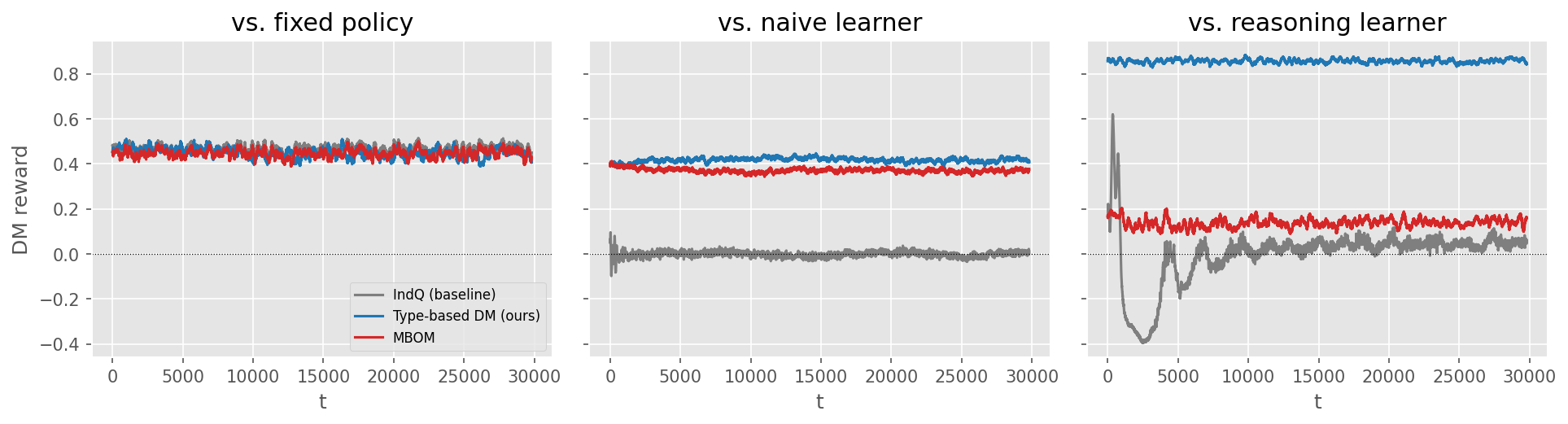}
\caption{\textcolor{black}{Reward of supported DM (equilibrium value $0$; higher is better) in iterated Matching Pennies against the three opponent classes in \citet{yu2022model}. Our type-based DM (blue) matches or exceeds MBOM (red) throughout and dominates against the reasoning learner, while opponent-unaware IndQ (gray) exploits none.}}
\label{fig:mbom}
\end{figure*}

\textcolor{black}{The comparisons above place the recent methods in the adversary's seat.
  We complement them by benchmarking, in the DM's seat, against MBOM \citep{yu2022model}, the recent method closest to our design: it builds a hierarchy of imagined opponent policies by recursive best response and mixes them through Bayesian similarity to observed behavior, a deep counterpart of the level-$k$ recursion and type-based averaging of Sections~\ref{sec:k} and~\ref{sec:com}. As this recursive imagination is informative only in competitive games (in Chicken it collapses onto a pure equilibrium), we adopt MBOM's own protocol: iterated Matching Pennies (the DM scores $+1$ when its action matches the opponent's, $-1$ otherwise; equilibrium value $0$) against the three opponent classes it considers, fixed, naive, and reasoning learner. Our DM is the type-based scheme of Section~\ref{sec:com}, spanning the same model range as MBOM's imagined policies (a non-strategic branch tracking action frequencies and a level-1 branch), selected online with a Bayesian posterior over types. Figure~\ref{fig:mbom} shows it matches MBOM against the fixed policy ($+0.45$), slightly exceeds it against the naive learner ($+0.42$ vs.\ $+0.37$), and dominates against the reasoning learner ($+0.86$ vs.\ $+0.14$), while an opponent-unaware IndQ exploits none. MBOM attains uniform but modest adaptation; the type-based DM instead recovers the best response against every class of opponents.}

\subsection{\textcolor{black}{Discussion}}\label{sec:discussion}

\textcolor{black}{We close the experimental section distilling the main experimental insights and practical implications.} 
Empirical evidence supporting TMDPs has been provided via extensive experiments with encouraging results. In security settings, we see that by explicitly modeling a finite set of adversaries via the opponent averaging scheme, a supported DM can take advantage of her actual opponent, even when he is not explicitly modeled through
one of the components in the finite mixture. This highlights the ability of our framework to generalize between different kinds of opponents. Interestingly, we find that a level-2 $Q$-learner may effectively deal with a wide class of adversaries. However, maintaining a mixture of different adversaries is necessary if we consider a level-3 DM. As a general heuristic rule, the supported DM may start at a low level in the hierarchy, and switch to a level-up temporarily, to check if the obtained rewards are higher. Otherwise, she would continue at the initial, lower level.

\textcolor{black}{Two practical properties of the proposed scheme are also delivered through 
 the experiments. First, it is robust, as it offers protection against adversarial behaviors that are not exactly as described by the DM's opponent model. Second, these benefits come at a reasonable cost: the increase in time and space complexity is linear compared to unprotected, baseline $Q$-learners.}

\textcolor{black}{Our experiments have compared against \textcolor{black}{seven} complementary baselines: (i) an independent (opponent-unaware) tabular $Q$-learner, which is the natural ablation, since our scheme is a drop-in modification of $Q$-learning; (ii) the DQN behavior reported by \citet{leike2017ai} on the Friend-or-Foe environment; (iii) adaptive exponential-smoother adversaries of varying sophistication; (iv) within-framework comparisons across cognitive levels, a non-trivial baseline since recursive-reasoning schemes such as \citet{wen2019probabilistic} only operate at level-2; (v) the WoLF-PHC \citep{bowling2001rational} comparison in Section~\ref{sec:wolf}\textcolor{black}{; (vi) the comparisons against LOQA \citep{aghajohari2024loqa} and POLA \citep{zhao2022proximal}, two state-of-the-art opponent-shaping methods, in the same section; and (vii) MBOM \citep{yu2022model}, a recent model-based recursive-reasoning method, benchmarked as an alternative decision-maker there}. We do not benchmark against the Markov-game family (minimax-$Q$ \citep{littman1994markov}, Nash-$Q$ \citep{hu2003nash}, friend-or-foe-$Q$ \citep{littman2001friend}) because they require common knowledge of payoffs, zero-sum, or known friend/foe status; LOLA \citep{foerster2018learning} \textcolor{black}{requires access to the opponent's policy parameters or gradients}; DPIQN/DRON \citep{he2016opponent} are tied to a specific NN architecture; PSRO \citep{lanctot2017unified} relies on a meta-game oracle; and \citet{pinto2017robust} address only zero-sum perturbations. \textcolor{black}{Analogous obstacles affect another recent generation of methods: M-FOS \citep{lu2022model} dispenses with opponent gradients but requires meta-training over many opponent lifetimes in a resettable simulator; LIAM \citep{papoudakis2021agent} and GSCU \citep{fu2022greedy} depend on pretrained pools of opponent policies.} None of these assumptions hold under the limited-information setting natural to ARA\textcolor{black}{, in which a single lifetime of play against an unknown opponent is available. The exceptions, all compared against in Section~\ref{sec:loqa}, are LOQA (whose requirements coincide with ours), POLA (run as a shaping adversary), and MBOM (whose environment model is available in closed form for the tabular games studied, so it serves as an alternative decision-maker)}. The motivation for selecting $Q$-learning, and the transfer to policy-based RL, was given at the end of Section~\ref{sec:background}.}

\textcolor{black}{A natural concern is that TMDPs assume the DM eventually observes her opponent's actions and rewards, 
 whereas in security and cybersecurity these signals can be partially observable. Two remarks. First, in the application domains we target, post-hoc observability is the rule rather than the exception: intrusion-detection logs, forensic traces, threat-intelligence feeds, and after-action reports routinely reveal attacker actions and the damage they caused, even when unavailable in real time. Second, the framework is modular: the opponent forecast $p_A(b\mid s)$ can be replaced by a posterior over latent opponent actions learned from noisy or delayed observations, without altering the level-$k$ machinery. The full-information POSG and i-POMDP formulations that drop this assumption altogether are intractable beyond toy problems (Section~\ref{sec:background}), so the stated assumption is a deliberate tractability trade-off.
  A full partial-observability extension and head-to-head comparison against truncated i-POMDPs constitutes promising future work.}

\textcolor{black}{A second natural question is how TMDPs compare to policy-based RL (PG, TRPO, PPO) when scaling to continuous or high-dimensional action spaces. As anticipated at the end of Section~\ref{sec:background}, the construction lifts to those settings essentially unchanged. Here we discuss the resulting trade-offs. $Q$-learning is off-policy and sample-efficient, admits a contractive Bellman operator (Appendix~A), and yields the linear-in-$k$ overheads reported in Section~\ref{sec:cc}. These properties fit the discrete, security-flavored decisions we target. Conversely, the $\max_a$ in $Q$-learning is awkward in continuous action spaces and a poor fit for mixed equilibria. In those regimes, stochastic policy-based methods handle continuous controls natively and benefit from the trust-region or clipped-surrogate machinery of TRPO and PPO for stable updates. The trade-off they buy is more acute in non-stationary multi-agent settings: policy gradients have higher variance, on-policy data demands, and weaker theory, and the opponent's drift further inflates the cost of fresh interactions.}

\section{Conclusions and Further Work}

\textcolor{black}{
We have introduced TMDPs, a reformulation of MDPs to 
\textcolor{black}{ support decision-makers who confront opponents interfering  }
with the reward generation process in RL settings.
\textcolor{black}{Potential applications abound. 
  As examples, we mention developing schemes
for pilot decision support in air combat modeling, see \citet{virtanen2006modeling} for motivation; for autonomous driving 
decision support, see \citet{naveiro2022adversarial} for a static version
of an example problem in lane-changing support;  or for dynamic pricing,
see a discussion in 
\citet{rasines2024personalized}.}}

TMDPs aim at providing one-sided prescriptive support to an RL agent, maximizing her
expected reward taking into account predictions about potential negative actions adopted by an adversary.
\textcolor{black}{  The learning rule (\ref{eq:lr})-(\ref{eq:lr2}) is a contraction mapping (Appendix A) 
 and we may use RL convergence results, while gaining advantage from
opponent modeling within $Q$-learning. Indeed, we have proposed a scheme to model adversarial behavior based on level-$k$ reasoning
 extended with type-based reasoning to account for uncertainty about the opponent's level. Finally, we have sketched how the framework could be extended to deep learning settings and multiple opponents.  Empirical insights and practical properties (generality, robustness, and complexity) were discussed in Section~\ref{sec:discussion}.}

\textcolor{black}{
Several lines of work are possible for further research. First of all, in the experiments, we have just considered DMs up to level-3,
though the extension to higher-order adversaries is relevant. 
In addition, \textcolor{black}{ a concrete next step is the policy-gradient instantiation of the framework for continuous-action and high-dimensional settings, which would also enable a model-agnostic extension to deep multi-agent RL scenarios such as those of \citet{mnih2015human}.}
 Finally, it would be interesting to explore similar expansions to semi-MDPs, in order to perform hierarchical RL or allow for time-dependent rewards and transitions between states.}

\textcolor{black}
{\small 
\paragraph{Acknowledgements}{This manuscript was initially prepared while
the authors were visiting SAMSI within the Games and Decisions in Risk and Reliability program, under NSF Grant DMS-1638521.
The authors also acknowledge support from
the Severo Ochoa Excellence Programme CEX-2023-001347-S, Ministry of Science programs PID2021-124662OB-I00 and PID2025-172412OB-C21, the European Union's Horizon 2020 Research and Innovation Programme under Grant Agreement No. 101021797 (STARLIGHT), the FBBVA AMALFI project, and the SEDIA Excellent AI project EMOROBCARE.
VG acknowledges support from grant FPU16-05034 and PTQ2021-011758,
RN acknowledges support from grant FPU15-03636. RN acknowledges the 2026 Leonardo Grant for Scientific Research and Cultural Creation from the BBVA Foundation. The BBVA Foundation accepts no responsibility for the opinions, comments, and content included in the project and/or the results derived from it, which are the sole and absolute responsibility of their authors.
DRI is grateful to the AXA-ICMAT Chair
in Adversarial Risk Analysis.
}
}

\begingroup
\footnotesize
\textcolor{black}{
\bibliography{mybibfile}

@article{powell2019unified,
  title={A unified framework for stochastic optimization},
  author={Powell, Warren B},
  journal={European Journal of Operational Research},
  volume={275},
  number={3},
  pages={795--821},
  year={2019},
  publisher={Elsevier},
  doi={10.1016/j.ejor.2018.07.014}
}

@book{murphy2023probabilistic,
  title={Probabilistic machine learning: Advanced topics},
  author={Murphy, Kevin P},
  year={2023},
  publisher={MIT press},
  isbn={9780262048439},
  note={{ISBN:} 9780262048439},
}

@ARTICLE{roponen,
	author = {Roponen, J. and Rios Insua, D. and Salo, A.},
	title = {Adversarial risk analysis under partial information},
	journal = {European Journal of Operational Research},
	year = {2020},
	volume = {287},
	pages = {306-316},
	number = {1},
	month = {November},
	publisher = {Elsevier},
	doi = {10.1016/j.ejor.2020.04.037}
}

@ARTICLE{leelee,
	author = {Lee, H. and Lee, T, T.},
	title = {Multi-agent reinforcement learning algorithm to solve a partially-observable multi-agent problem in disaster response},
		journal = {Eur. Jour. Oper. Res.},
	year = {2021},
	volume = {291},
	pages = {296-308},
	number = {1},
	month = {November},
	publisher = {Elsevier},
	doi = {10.1016/j.ejor.2020.09.018}
}

@article{melo2001convergence,
  title={Convergence of Q-learning: A simple proof},
  author={Melo, Francisco S},
  journal={Tech. Rep.},
  year={2001},
  url={http://users.isr.ist.utl.pt/~mtjspaan/readingGroup/ProofQlearning.pdf}
}

@article{brown1951iterative,
  title={Iterative Solution of Games by Fictitious Play},
  author={Brown, George W},
  journal={Activity Analysis of Production and Allocation},
  pages={374--376},
  year={1951},
  publisher={Wiley}
}

@article{rios2012adversarial,
  title={Adversarial risk analysis for counterterrorism modeling},
  author={Rios, Jesus and Insua, David },
  journal={Risk Analysis: An International Journal},
  volume={32},
  number={5},
  pages={894--915},
  year={2012},
  publisher={Wiley Online Library},
  doi={10.1111/j.1539-6924.2011.01713.x}
}

@article{Albrecht2018AutonomousAM,
  title={Autonomous agents modelling other agents: A comprehensive survey and open problems},
  author={Stefano V. Albrecht and Peter Stone},
  journal={Artif. Intell.},
  year={2018},
  volume={258},
  pages={66-95},
  doi={10.1016/j.artint.2018.01.002}
}

@inproceedings{bowling2001rational,
  title={Rational and convergent learning in stochastic games},
  author={Bowling, Michael and Veloso, Manuela},
  booktitle={Proc. 17th Int. Joint Conf. AI},
  pages={1021--1026},
  year={2001},
  organization={Morgan Kaufmann.},
  url={https://dl.acm.org/doi/10.5555/1642194.1642231}
}

@article{raftery1985model,
  title={A model for high-order Markov chains},
  author={Raftery, Adrian E},
  journal={Journal of the Royal Statistical Society. Series B (Methodological)},
  pages={528--539},
  year={1985},
  publisher={JSTOR},
  doi={10.1111/j.2517-6161.1985.tb01383.x}
}

@article{lin2017tactics,
  title={Tactics of adversarial attack on deep reinforcement learning agents},
  author={Lin, Yen-Chen and Hong, Zhang-Wei and Liao, Yuan-Hong and Shih, Meng-Li and Liu, Ming-Yu and Sun, Min},
  journal={arXiv preprint arXiv:1703.06748},
  year={2017},
  doi={10.24963/ijcai.2017/525}
}

@article{leike2017ai,
  title={{AI} Safety Gridworlds},
  author={Leike, Jan and Martic, Miljan and Krakovna, Victoria and Ortega, Pedro A and Everitt, Tom and Lefrancq, Andrew and Orseau, Laurent and Legg, Shane},
  journal={arXiv preprint arXiv:1711.09883},
  year={2017},
  doi={10.48550/arXiv.1711.09883}
}

@article{rios1,
author = {Insua, David and David Banks and Jesus Rios},
title = {Modeling Opponents in Adversarial Risk Analysis},
journal = {Risk Analysis},
volume = {36},
number = {4},
pages = {742-755},
year = {2016},
doi = {10.1111/risa.12439}
}

@inproceedings{baxter2000direct,
  title={Direct gradient-based reinforcement learning},
  author={Baxter, Jonathan and Bartlett, Peter L},
  booktitle={2000 IEEE Int. Symp. Circ. Sys. Proc. },
  volume={3},
  pages={271--274},
  year={2000},
  organization={IEEE},
  doi={10.1109/ISCAS.2000.856049}
}

@inproceedings{tang2017exploration,
  title={\# Exploration: A Study of Count-Based Exploration for Deep Reinforcement Learning},
  author={Tang, Haoran and Houthooft, Rein and Foote, Davis and Stooke, Adam and Chen, OpenAI Xi and Duan, Yan and Schulman, John and DeTurck, Filip and Abbeel, Pieter},
  booktitle={Advances in Neural Information Processing Systems},
  pages={2750--2759},
  year={2017},
  doi={10.48550/arXiv.1611.04717}
}

@article{banks2022adversarial,
  title={Adversarial risk analysis: An overview},
  author={Banks, David and Gallego, V{\'\i}ctor and Naveiro, Roi and R{\'\i}os Insua, David},
  journal={Wiley Interdisciplinary Reviews: Computational Statistics},
  volume={14},
  number={1},
  pages={e1530},
  year={2022},
  publisher={Wiley Online Library},
  doi={10.1002/wics.1530}
}

@inproceedings{gallego2019reinforcement,
  title={Reinforcement learning under threats},
  author={Gallego, Victor and Naveiro, Roi and Insua, David Rios},
  booktitle={Proceedings of the AAAI Conference on Artificial Intelligence},
  volume={33},
  number={01},
  pages={9939--9940},
  year={2019},
  doi={10.1609/aaai.v33i01.33019939}
}

@article{rios2009adversarial,
  title={Adversarial risk analysis},
  author={Insua, David and Rios, Jesus and Banks, David},
  journal={Journal of the American Statistical Association},
  volume={104},
  number={486},
  pages={841--854},
  year={2009},
  publisher={Taylor \& Francis},
  doi={10.1198/jasa.2009.0155}
}

@article{mnih2015human,
  title={Human-level control through deep reinforcement learning},
  author={Mnih, Volodymyr and Kavukcuoglu, Koray and Silver, David and Rusu, Andrei A and Veness, Joel and Bellemare, Marc G and Graves, Alex and Riedmiller, Martin and Fidjeland, Andreas K and Ostrovski, Georg and others},
  journal={Nature},
  volume={518},
  number={7540},
  pages={529},
  year={2015},
  publisher={Nature Publishing Group},
  doi={10.1038/nature14236}
}

@article{silver2017mastering,
  title={Mastering the game of go without human knowledge},
  author={Silver, David and Schrittwieser, Julian and Simonyan, Karen and Antonoglou, Ioannis and Huang, Aja and Guez, Arthur and Hubert, Thomas and Baker, Lucas and Lai, Matthew and Bolton, Adrian and others},
  journal={Nature},
  volume={550},
  number={7676},
  pages={354},
  year={2017},
  publisher={Nature Publishing Group},
  doi={10.1038/nature24270}
}

@inproceedings{he2016opponent,
  title={Opponent modeling in deep reinforcement learning},
  author={He, He and Boyd-Graber, Jordan and Kwok, Kevin and Daume III, Hal},
  booktitle={International Conference on Machine Learning},
  pages={1804--1813},
  year={2016},
  doi={10.48550/arXiv.1609.05559}
}

@inproceedings{foerster2018learning,
  title={Learning with opponent-learning awareness},
  author={Foerster, Jakob and Chen, Richard Y and Al-Shedivat, Maruan and Whiteson, Shimon and Abbeel, Pieter and Mordatch, Igor},
  booktitle={Proc. 17th Int. Conf. Aut. Agents and MultiAgent Systems},
  pages={122--130},
  year={2018},
  url={https://dl.acm.org/doi/10.5555/3237383.3237408}
}

@incollection{littman1994markov,
  title={Markov games as a framework for multi-agent reinforcement learning},
  author={Littman, Michael L},
  booktitle={Machine Learning Proceedings 1994},
  pages={157--163},
  year={1994},
  publisher={Elsevier},
  doi={10.1016/B978-1-55860-335-6.50027-1}
}

@article{hu2003nash,
  title={{Nash Q-learning for general-sum stochastic games}},
  author={Hu, Junling and Wellman, Michael P},
  journal={Journal of Machine Learning Research},
  volume={4},
  number={Nov},
  pages={1039--1069},
  year={2003},
  url={https://dl.acm.org/doi/10.5555/945365.964288}
}

@inproceedings{littman2001friend,
  title={{Friend-or-Foe Q-learning in General-Sum Games}},
  author={Littman, Michael L},
  booktitle={Proc. of the 18th ICML},
  pages={322--328},
  year={2001},
  url={https://dl.acm.org/doi/10.5555/645530.655661}
}

@inproceedings{auer1995gambling,
  title={Gambling in a rigged casino: The adversarial multi-armed bandit problem},
  author={Auer, Peter and Cesa-Bianchi, Nicolo and Freund, Yoav and Schapire, Robert E},
  booktitle={Proc. 36th Symp. Found. CS.},
  pages={322--331},
  year={1995},
  doi={10.1109/SFCS.1995.492488}
}

@book{howard:dp,
  address = {Cambridge, MA},
  author = {Howard, R. A.},
  publisher = {MIT Press},
  title = {Dynamic Programming and Markov Processes},
  year = 1960,
  isbn = {9780262080095}
}

@book{powell2022reinforcement,
  title={Reinforcement Learning and Stochastic Optimization: A Unified Framework for Stochastic Optimization},
  author={Powell, Warren B.},
  year={2022},
  publisher={John Wiley \& Sons},
  isbn={9781119815037},
  doi={10.1002/9781119815068}
}

@article{stahl1994experimental,
  title={Experimental evidence on players' models of other players},
  author={Stahl, Dale O and Wilson, Paul W},
  journal={Journal of Economic Behavior \& Organization},
  volume={25},
  number={3},
  pages={309--327},
  year={1994},
  publisher={Elsevier},
  doi={10.1016/0167-2681(94)90103-1}
}

@inproceedings{dalvi2004adversarial,
  title={Adversarial classification},
  author={Dalvi, Nilesh and Domingos, Pedro and Sanghai, Sumit and Verma, Deepak and others},
  booktitle={Proc. 10th ACM SIGKDD},
  pages={99--108},
  year={2004},
  organization={ACM},
  doi={10.1145/1014052.1014066}
}

@article{hart2008discrete,
  title={{Discrete Colonel Blotto and General Lotto Games}},
  author={Hart, Sergiu},
  journal={International Journal of Game Theory},
  volume={36},
  number={3-4},
  pages={441--460},
  year={2008},
  publisher={Springer},
  doi={10.1007/s00182-007-0099-9}
}

@article{menache2011network,
  title={Network games: Theory, models, and dynamics},
  author={Menache, Ishai and Ozdaglar, Asuman},
  journal={Synthesis Lectures on Communication Networks},
  volume={4},
  number={1},
  pages={1--159},
  year={2011},
  publisher={Morgan \& Claypool Publishers},
  doi={10.2200/S00330ED1V01Y201101CNT009}
}

@article{BIGGIO2018317,
title = "Wild patterns: Ten years after the rise of adversarial machine learning",
journal = "Pattern Recognition",
volume = "84",
pages = "317 - 331",
year = "2018",
issn = "0031-3203",
author = "Battista Biggio and Fabio Roli",
doi = "10.1016/j.patcog.2018.07.023"
}

@inproceedings{behzadan2017vulnerability,
  title={Vulnerability of deep reinforcement learning to policy induction attacks},
  author={Behzadan, Vahid and Munir, Arslan},
  booktitle={Mach. Learn. Data Mining in Patt. Recog. 13th Int. Conf., MLDM},
  year={2017},
  organization={Springer},
  doi={10.1007/978-3-319-62416-7_19}
}

@book{sutton2012reinforcement,
  title={Reinforcement learning: An introduction},
  author={Sutton, Richard S and Barto, Andrew G},
  year={2018},
  publisher={MIT Press},
  address={Cambridge, MA},
  isbn={9780262039246},
  note={{ISBN:} 9780262039246}
}

@InProceedings{pmlr-v37-schulman15,
  title = 	 {Trust Region Policy Optimization},
  author = 	 {Schulman, John and Levine, Sergey and Abbeel, Pieter and Jordan, Michael and Moritz, Philipp},
  booktitle = 	 {Proceedings of the 32nd International Conference on Machine Learning},
  pages = 	 {1889--1897},
  year = 	 {2015},
  editor = 	 {Bach, Francis and Blei, David},
  volume = 	 {37},
  series = 	 {Proceedings of Machine Learning Research},
  address = 	 {Lille, France},
  month = 	 {07--09 Jul},
  publisher =    {PMLR},
  doi = {10.48550/arXiv.1502.05477}
}

@inproceedings{camacho2024algorithmic,
  title={Algorithmic Decision Analysis for Multi-stage Games with Incomplete Information},
  author={Camacho, Jos{\'e} Manuel and Naveiro, Roi and R{\'\i}os Insua, David},
  booktitle={Int. Conf. Alg. Dec. Theory},
  pages={98--112},
  year={2024},
  organization={Springer},
  doi={10.1007/978-3-031-73903-3_7}
}

@article{schulman2017proximal,
  title={Proximal policy optimization algorithms},
  author={Schulman, John and Wolski, Filip and Dhariwal, Prafulla and Radford, Alec and Klimov, Oleg},
  journal={arXiv preprint arXiv:1707.06347},
  year={2017},
  doi={10.48550/arXiv.1707.06347}
}

@article{kadane1982subjective,
  title={Subjective probability and the theory of games},
  author={Kadane, Joseph B and Larkey, Patrick D},
  journal={Management Science},
  volume={28},
  number={2},
  pages={113--120},
  year={1982},
  publisher={INFORMS},
  doi={10.1287/mnsc.28.2.113}
}

@book{raiffa1982art,
  title={The Art and Science of Negotiation},
  author={Raiffa, H.},
  isbn={9780674048133},
  lccn={82006170},
  year={1982},
  publisher={Belknap Press of Harvard University Press},
  address={Cambridge, MA},
  note={{ISBN:} 9780674048133},
}

@article{naveiro2018adversarial,
  title={Adversarial classification: An adversarial risk analysis approach},
  author={Naveiro, Roi and Redondo, Alberto and Insua, David  and Ruggeri, Fabrizio},
  journal={Int. Journal of Approximate Reasoning},
  year={2019},
  publisher={Elsevier},
  doi={10.1016/j.ijar.2019.07.003}
}

@article{camerer2004cognitive,
  title={A cognitive hierarchy model of games},
  author={Camerer, Colin F and Ho, Teck-Hua and Chong, Juin-Kuan},
  journal={The Quarterly Journal of Economics},
  volume={119},
  number={3},
  pages={861--898},
  year={2004},
  publisher={MIT Press},
  doi={10.1162/0033553041502225}
}

@article{doi:10.1002/widm.1259,
author = {Zhou, Yan and Kantarcioglu, Murat and Xi, Bowei},
title = {A survey of game theoretic approach for adversarial machine learning},
journal = {Wiley Interdisciplinary Reviews: Data Mining and Knowledge Discovery},
year={2018},
doi = {10.1002/widm.1259},
}

@book{hargreaves2004game,
  title={Game Theory: A Critical Introduction},
  author={Hargreaves-Heap, S. and Varoufakis, Y.},
  isbn={9780203489291},
  lccn={2003058677},
  year={1995},
  publisher={Routledge},
}

@article{huang2017adversarial,
  title={Adversarial attacks on neural network policies},
  author={Huang, Sandy and Papernot, Nicolas and Goodfellow, Ian and Duan, Yan and Abbeel, Pieter},
  journal={arXiv preprint arXiv:1702.02284},
  year={2017},
  doi={10.48550/arXiv.1702.02284}
}

@article{goodfellow2014explaining,
  title={Explaining and Harnessing Adversarial Examples},
  author={Goodfellow, Ian J and Shlens, Jonathon and Szegedy, Christian},
  journal={arXiv preprint arXiv:1412.6572},
  year={2014},
  doi={10.48550/arXiv.1412.6572}
}

@article{jasa,
author = {David Rios Insua and Roi Naveiro and Víctor Gallego and Jason Poulos},
title = {Adversarial Machine Learning: Bayesian Perspectives},
journal = {Journal of the American Statistical Association},
volume = {0},
number = {0},
pages = {1-12},
year  = {2023},
publisher = {Taylor & Francis},
doi = {10.1080/01621459.2023.2183129}
}

@article{virtanen2006modeling,
  title={Modeling air combat by a moving horizon influence diagram game},
  author={Virtanen, Kai and Karelahti, Janne and Raivio, Tuomas},
  journal={Journal of guidance, control, and dynamics},
  volume={29},
  number={5},
  pages={1080--1091},
  year={2006},
  doi={10.2514/1.17168}
}

@book{DEEPLEARNING,
  title={Deep Learning},
  author={Goodfellow, Ian and Bengio, Y. and Courville, A.},
    year={2016},
  publisher={MIT Press},
  isbn={9780262035613},
  note={{ISBN:} 9780262035613}
}

@article{gallego2022current,
  title={Current advances in neural networks},
  author={Gallego, V{\'\i}ctor and R{\'\i}os Insua, David},
  journal={Annual Review of Statistics and Its Application},
  volume={9},
  number={1},
  pages={197--222},
  year={2022},
  publisher={Annual Reviews},
  doi={10.1146/annurev-statistics-040220-112019}
}

@article{gallego2024protecting,
  title={Protecting classifiers from attacks},
  author={Gallego, Victor and Naveiro, Roi and Redondo, Alberto and R{\'\i}os Insua, David and Ruggeri, Fabrizio},
  journal={Statistical Science},
  volume={39},
  number={3},
  pages={449--468},
  year={2024},
  publisher={Institute of Mathematical Statistics}
}

@article{banihashem2021defense,
  title={Defense against reward poisoning attacks in reinforcement learning},
  author={Banihashem, Kiarash and Singla, Adish and Radanovic, Goran},
  journal={arXiv preprint arXiv:2102.05776},
  year={2021},
  doi={10.48550/arXiv.2102.05776}
}

@inproceedings{pinto2017robust,
  title={Robust Adversarial Reinforcement Learning},
  author={Pinto, Lerrel and Davidson, James and Sukthankar, Rahul and Gupta, Abhinav},
  booktitle={International Conference on Machine Learning},
  pages={2817--2826},
  year={2017},
  doi={10.48550/arXiv.1703.02702}
}

@inproceedings{wen2019probabilistic,
  title={Probabilistic recursive reasoning for multi-agent reinforcement learning},
  author={Wen, Y and Yang, Y and Luo, R and Wang, J and Pan, W},
  booktitle={7th International Conference on Learning Representations},
  volume={7},
  year={2019},
  doi={10.48550/arXiv.1901.09207}
}

@article{tretschk2018sequential,
  title={Sequential attacks on agents for long-term adversarial goals},
  author={Tretschk, Edgar and Oh, Seong Joon and Fritz, Mario},
  journal={arXiv preprint arXiv:1805.12487},
  year={2018},
  doi={10.48550/arXiv.1805.12487}
}

@article{chen2019adversarial,
  title={Adversarial attack and defense in reinforcement learning-from AI security view},
  author={Chen, Tong and Liu, Jiqiang and Xiang, Yingxiao and Niu, Wenjia and Tong, Endong and Han, Zhen},
  journal={Cybersecurity},
  volume={2},
  pages={1--22},
  year={2019},
  publisher={Springer},
  doi={10.1186/s42400-019-0027-x}
}

@article{zhang2020robust,
  title={Robust deep reinforcement learning against adversarial perturbations on state observations},
  author={Zhang, Huan and Chen, Hongge and Xiao, Chaowei and Li, Bo and Liu, Mingyan and Boning, Duane and Hsieh, Cho-Jui},
  journal={Advances in Neural Information Processing Systems},
  volume={33},
  pages={21024--21037},
  year={2020},
  doi={10.48550/arXiv.2003.08938}
}

@article{caballero2021poisoning,
  title={Poisoning finite-horizon Markov decision processes at design time},
  author={Caballero, William N and Jenkins, Phillip R and Keith, Andrew J},
  journal={Computers \& Operations Research},
  volume={129},
  pages={105185},
  year={2021},
  publisher={Elsevier},
  doi={10.1016/j.cor.2020.105185}
}

@article{caballero2021identifying,
  title={Identifying behaviorally robust strategies for normal form games under varying forms of uncertainty},
  author={Caballero, William N and Lunday, Brian J and Uber, Richard P},
  journal={Eur. Jour. Oper. Res.},
  volume={288},
  number={3},
  year={2021},
  pages={971--982},
  publisher={Elsevier},
  doi={10.1016/j.ejor.2020.06.022}
}

@incollection{marl_over,
  title={Multi-agent reinforcement learning: An overview},
  author={Busoniu, Lucian and Babuska, Robert and De Schutter, Bart},
  booktitle={Innovations in Multi-Agent Systems and Applications-1},
  pages={183--221},
  year={2010},
  publisher={Springer},
  doi={10.1007/978-3-642-14435-6_7}
}

@ARTICLE{BusoniuMARL,
  author={Busoniu, Lucian and Babuska, Robert and De Schutter, Bart},
  journal={IEEE Transactions on Systems, Man, and Cybernetics, Part C (Applications and Reviews)}, 
  title={A Comprehensive Survey of Multiagent Reinforcement Learning}, 
  year={2008},
  volume={38},
  number={2},
  pages={156-172},
  doi={10.1109/TSMCC.2007.913919}}

@inproceedings{lanctot2017unified,
  title={A unified game-theoretic approach to multiagent reinforcement learning},
  author={Lanctot, Marc and Zambaldi, Vinicius and Gruslys, Audrunas and Lazaridou, Angeliki and Tuyls, Karl and Perolat, Julien and Silver, David and Graepel, Thore},
  booktitle={Advances in Neural Information Processing Systems},
  pages={4190--4203},
  year={2017},
  doi={10.48550/arXiv.1711.00832}
}

@article{heinrich2016deep,
  title={Deep reinforcement learning from self-play in imperfect-information games},
  author={Heinrich, Johannes and Silver, David},
  journal={arXiv preprint arXiv:1603.01121},
  year={2016},
  doi={10.48550/arXiv.1603.01121}
}

@inproceedings{hansen2004dynamic,
  title={Dynamic programming for partially observable stochastic games},
  author={Hansen, Eric A. and Bernstein, Daniel S. and Zilberstein, Shlomo},
  booktitle={AAAI Conference on Artificial Intelligence},
  pages={709--715},
  year={2004},
  url={https://dl.acm.org/doi/10.5555/1597148.1597262}
}

@article{konda2000actor,
  author = {Konda, Vijay and Tsitsiklis, John},
 booktitle = {Advances in Neural Information Processing Systems},
 editor = {S. Solla and T. Leen and K. M\"{u}ller},
 pages = {},
 publisher = {MIT Press},
 title = {Actor-Critic Algorithms},
 url = {https://dl.acm.org/doi/10.5555/3009657.3009799},
 volume = {12},
 year = {1999}
}

@article{gmytrasiewicz2005framework,
  title={A framework for sequential planning in multi-agent settings},
  author={Gmytrasiewicz, Piotr J and Doshi, Prashant},
  journal={Journal of Artificial Intelligence Research},
  volume={24},
  pages={49--79},
  year={2005},
  doi={10.1613/jair.1579}
}

@article{kaelbling1996reinforcement,
  title={Reinforcement learning: A survey},
  author={Kaelbling, Leslie Pack and Littman, Michael L and Moore, Andrew W},
  journal={Journal of Artificial Intelligence Research},
  volume={4},
  pages={237--285},
  year={1996},
  doi={10.1613/jair.301}
}

@article{rasines2024personalized,
  title={Personalized pricing decisions through adversarial risk analysis},
  author={Rasines, Daniel Garc{\'\i}a and Naveiro, Roi and Insua, David R{\'\i}os and Santana, Sim{\'o}n Rodr{\'\i}guez},
  journal={International Transactions in Operational Research},
  year={2024},
  publisher={Wiley Online Library},
  doi={10.1111/itor.13545}
}

@article{naveiro2022adversarial,
  title={Adversarial Risk Analysis for Heterogeneous Traffic Management},
  author={Naveiro, Roi and Caballero, William N and R{\'\i}os Insua, David},
  journal={Available at SSRN 4365987},
  year={2022},
  doi={10.2139/ssrn.4365987}
}

@inproceedings{aghajohari2024loqa,
  title={{LOQA}: Learning with opponent {Q}-learning awareness},
  author={Aghajohari, Milad and Duque, Juan Agustin and Cooijmans, Tim and Courville, Aaron},
  booktitle={12th International Conference on Learning Representations},
  year={2024},
  doi={10.48550/arXiv.2405.01035}
}

@inproceedings{lu2022model,
  title={Model-free opponent shaping},
  author={Lu, Chris and Willi, Timon and Schroeder de Witt, Christian and Foerster, Jakob},
  booktitle={Proceedings of the 39th International Conference on Machine Learning},
  pages={14398--14411},
  year={2022}
}

@inproceedings{zhao2022proximal,
  title={Proximal learning with opponent-learning awareness},
  author={Zhao, Stephen and Lu, Chris and Grosse, Roger B and Foerster, Jakob N},
  booktitle={Advances in Neural Information Processing Systems},
  volume={35},
  year={2022},
  doi={10.48550/arXiv.2210.10125}
}

@inproceedings{yu2022model,
  title={Model-based opponent modeling},
  author={Yu, Xiaopeng and Jiang, Jiechuan and Zhang, Wanpeng and Jiang, Haobin and Lu, Zongqing},
  booktitle={Advances in Neural Information Processing Systems},
  volume={35},
  year={2022},
  doi={10.48550/arXiv.2108.01843}
}

@inproceedings{papoudakis2021agent,
  title={Agent modelling under partial observability for deep reinforcement learning},
  author={Papoudakis, Georgios and Christianos, Filippos and Albrecht, Stefano V},
  booktitle={Advances in Neural Information Processing Systems},
  volume={34},
  year={2021},
  doi={10.48550/arXiv.2006.09447}
}

@inproceedings{fu2022greedy,
  title={Greedy when sure and conservative when uncertain about the opponents},
  author={Fu, Haobo and Tian, Ye and Yu, Hongxiang and Liu, Weiming and Wu, Shuang and Xiong, Jiechao and Wen, Ying and Li, Kai and Xing, Junliang and Fu, Qiang and Yang, Wei},
  booktitle={Proceedings of the 39th International Conference on Machine Learning},
  pages={6829--6848},
  year={2022}
}

@inproceedings{duque2025advantage,
  title={Advantage alignment algorithms},
  author={Duque, Juan Agustin and Aghajohari, Milad and Cooijmans, Tim and Ciuca, Razvan and Zhang, Tianyu and Gidel, Gauthier and Courville, Aaron},
  booktitle={13th International Conference on Learning Representations},
  year={2025},
  doi={10.48550/arXiv.2406.14662}
}

@article{keith2021counterfactual,
  title={Counterfactual regret minimization for integrated cyber and air defense resource allocation},
  author={Keith, Andrew J. and Ahner, Darryl K.},
  journal={European Journal of Operational Research},
  volume={292},
  number={1},
  pages={95--107},
  year={2021},
  publisher={Elsevier},
  doi={10.1016/j.ejor.2020.10.015}
}

@article{boute2022deep,
  title={Deep reinforcement learning for inventory control: A roadmap},
  author={Boute, Robert N. and Gijsbrechts, Joren and van Jaarsveld, Willem and Vanvuchelen, Nathalie},
  journal={European Journal of Operational Research},
  volume={298},
  number={2},
  pages={401--412},
  year={2022},
  publisher={Elsevier},
  doi={10.1016/j.ejor.2021.07.016}
}

@article{hausken2024fifty,
  title={Fifty years of operations research in defense},
  author={Hausken, Kjell},
  journal={European Journal of Operational Research},
  volume={318},
  number={2},
  pages={355--368},
  year={2024},
  publisher={Elsevier}
}
}
\endgroup
\appendix
\section{Convergence proof of update rule for TMDPs}\label{sec:p}
\textcolor{black}{Given a TMDP, consider an augmented state space with transitions $(s,b) \xrightarrow[]{a} (s',b') \xrightarrow[]{a'} \ldots$. The DM does not observe the full state, since she does not know the action $b$ taken by her adversary, but it suffices that she knows his policy $p(b|s)$ or a good estimate of it. Assume the opponent's action $b$ is observed. Using the standard $Q$-learning recursion \citep{sutton2012reinforcement} and the conditional independence $p(s', b' |s,a,b) = p(b'|s')p(s'|s,a,b)$, the optimal $Q$-function satisfies
$$
Q^*(s,a,b) =  \sum_{s'} p(s'|s,a,b) \left[ R_{ss'}^{ab} + \gamma \max_{a'} \mathbb{E}_{p(b'|s')} \left[ Q^*(s',a',b') \right] \right],
$$
with $R_{ss'}^{ab} = \mathbb{E}\left[ r_{t+1}|s_{t+1} = s', s_{t} = s, a_t = a, b_t = b \right]$. Observe now that}
\textcolor{black}{
\begin{lemma}\label{lema:1}
Given $q: \mathcal{S} \times \mathcal{B} \times \mathcal{A} \rightarrow \mathbb{R}$,
the operator $\mathcal{H}$ 
\begin{flalign*}
(\mathcal{H}q) (s,b,a) = \sum_{s'} p(s'|s,b,a) \big[ r(s,b,a) +\gamma \max_{a'} \mathbb{E}_{p(b'|s')} q(s',b',a') \big]
\end{flalign*}
is a contraction mapping under the supremum norm. 
\end{lemma}
\begin{proof}
  Let us prove that $\| \mathcal{H}q_1 - \mathcal{H}q_2 \|_{\infty} \leq \gamma \| q_1 - q_2 \|_{\infty}$.
{\small\setlength{\jot}{2pt}
\begin{flalign*}
& \| \mathcal{H}q_1 - \mathcal{H}q_2 \|_{\infty} = \gamma \max_{s,b,a}   \Big\vert \sum_{s'} p(s'|s,b,a) \big[   \max_{a'} \mathbb{E}_{p(b'|s')} q_1(s',b',a')  -\max_{a'} \mathbb{E}_{p(b'|s')} q_2(s',b',a')\big]  \Big\vert \leq \\
&\leq \gamma \max_{s,b,a} \sum_{s'} p(s'|s,b,a)  \max_{a',z}  \Big\vert \mathbb{E}_{p(b'|z)} q_1(z,b',a') - \mathbb{E}_{p(b'|z)} q_2(z,b',a')  \Big\vert \leq \\
&\leq \gamma \max_{s,b,a} \sum_{s'} p(s'|s,b,a)  \max_{a',z,b'}  \Big\vert q_1(z,b',a') -  q_2(z,b',a')  \Big\vert = \gamma \|  q_1 -  q_2 \|_{\infty}.\hspace{2cm}\qedhere
\end{flalign*}}
\end{proof}
}

\noindent 
\textcolor{black}{
 Then, using the proposed learning rule (\ref{eq:lr}), we would converge
to the optimal $Q$ for each of the opponent actions. The result follows directly from the standard $Q$-learning convergence proof, see, e.g., {\color{black}\citet{melo2001convergence}}, using Lemma 1.}

\textcolor{black}{
However, at the time of making the {\color{black}decision we} do not know what action he would take. Thus, we average over the possible opponent
actions, weighting each of the actions by $p(b|s)$, as \eqref{eq:lr2} suggests.}

\color{black}
\section{Experiment details}\label{sec:times}
Experiments were run on a 16-core Intel i7-5960X server with 128\,GB RAM; code is at \url{https://github.com/vicgalle/ARAMARL}. Table~\ref{tab:times_combined} reports per-episode times, whose overhead grows linearly with the DM's cognitive level, consistent with Section~\ref{sec:cc}.

\begin{table}[H]
\centering
\footnotesize
\setlength{\tabcolsep}{4pt}
\caption{Experiment times. Sec.\ 4.1.1--4.1.2: 20 seeds, 5000 steps. Sec.\ 4.1.3 and 4.3: 5 seeds, 10000 steps.}
\label{tab:times_combined}
\begin{tabular}{lc|lc|lc}
\hline
\multicolumn{2}{c|}{\textbf{Sec.\ 4.1.1--4.1.2}} & \multicolumn{2}{c|}{\textbf{Sec.\ 4.1.3}} & \multicolumn{2}{c}{\textbf{Sec.\ 4.3}} \\
\hline
IndQ & 10.9\,s    & IndQ          & 3m 24s & RegressionIndQ          & 4m 23s \\
FPQwForget (L1) & 18.6\,s & Level2Q  & 5m 18s & RegressionLevel2Q       & 8m 29s \\
Level2Q & 25.7\,s &              &        &                          &        \\
Level3Q & 34.5\,s &              &        &                          &        \\
\hline
\end{tabular}
\end{table}

\end{document}